\documentclass{article}

\usepackage[numbers,sort&compress]{natbib}

\usepackage{arxiv}

\usepackage[utf8]{inputenc} % allow utf-8 input
\usepackage[T1]{fontenc}    % use 8-bit T1 fonts
\usepackage{hyperref}       % hyperlinks
\usepackage{url}            % simple URL typesetting
\usepackage{booktabs}       % professional-quality tables
\usepackage{amsmath}        % align, bmatrix, equation
\usepackage{algorithm}      % algorithm float
\usepackage{algorithmic}    % algorithmic environment
\usepackage{amsfonts}       % blackboard math symbols
\usepackage{nicefrac}       % compact symbols for 1/2, etc.
\usepackage{microtype}      % microtypography
\usepackage{xcolor}         % colors
\usepackage{graphicx}       % figures / \resizebox
\usepackage{multirow}       % \multirow cells in tables
\usepackage{doi}

\title{ReQuant: Fixed-Grid Discrete Refinement for Post-Training Quantization}

\author{%
  \bfseries
  Yongge Ma$^{1}$,
  Guoan Wang$^{2}$,
  Feiyu Wang$^{1}$,
  Yaoming Li$^{1}$,
  Qian Zhang$^{3}$,
  Zihan Yan$^{4}$,
  Yinjun Han$^{5}$,
  Tong Yang$^{1}$\thanks{Corresponding author: \texttt{yangtong@pku.edu.cn}}\\[0.7em]
  \normalfont\mdseries\small
  $^{1}$School of Computer Science, Peking University \\
  $^{2}$School of Software and Microelectronics, Peking University \\
  $^{3}$School of Physics, Peking University \\
  $^{4}$The Chinese University of Hong Kong, Shenzhen \\
  $^{5}$Central Research Institute, ZTE Corporation
}

\date{}

\renewcommand{\headeright}{A Preprint}
\renewcommand{\undertitle}{}
\renewcommand{\shorttitle}{\textbf{ReQuant: Fixed-Grid Discrete Refinement for Post-Training Quantization}}

\makeatletter
\renewcommand{\@maketitle}{%
  \vbox{%
    \hsize\textwidth
    \linewidth\hsize
    \vskip 0.1in
    \@toptitlebar
    \centering
    {\LARGE\bfseries\scshape \@title\par}
    \@bottomtitlebar
    \vskip 0.18in
    {\centering\normalfont\@author\par}
    \vskip 0.28in
  }
}
\makeatother

\hypersetup{
  pdftitle={ReQuant: Fixed-Grid Discrete Refinement for Post-Training Quantization},
  pdfauthor={Yongge Ma, Guoan Wang, Feiyu Wang, Yaoming Li, Qian Zhang, Zihan Yan, Yinjun Han, Tong Yang},
  pdfsubject={cs.LG, cs.CL},
  pdfkeywords={post-training quantization, large language models, model quantization, discrete optimization},
}

\begin{document}

\maketitle

\begin{abstract}
    Post-training quantization (PTQ) is widely used to reduce the memory and computational cost of large language models. Existing PTQ methods typically obtain an initial quantized model through heuristic rules or greedy optimization, and once quantization is completed the resulting integer assignments are usually treated as final. This observation motivates a complementary optimization stage \emph{within} PTQ that keeps quantized weights improvable after an executable quantized model has been produced, while preserving the quantized format. We introduce ReQuant, a backpropagation-free fixed-grid refinement procedure for this stage. Agnostic to the PTQ initializer, ReQuant takes an existing quantized model as a feasible starting point and iteratively revisits its discrete weight assignments on the fixed quantization grid. Accepted updates strictly reduce the mean squared reconstruction error and remain on the original grid. In this way, ReQuant turns the initially fixed PTQ output into an iteratively optimizable discrete solution and serves as a plug-and-play post-processing stage for existing PTQ pipelines. Experiments across diverse model families, bit-widths, and downstream tasks show that ReQuant consistently improves quantized models from heterogeneous PTQ initializers, with especially large gains on simple initializers and lower bit-widths. Notably, ReQuant can refine a simple round-to-nearest initialization across multiple sweeps until it approaches or surpasses GPTAQ under the same quantization format. These results establish ReQuant as a practical complementary stage for further improving existing PTQ pipelines.
\end{abstract}


\section{Introduction}
Large language models (LLMs) based on the transformer architecture have achieved remarkable performance across a wide range of tasks~\cite{vaswani2017attention,brown2020gpt3,achiam2023gpt4,liu2024deepseek,bai2023qwen}.
However, the rapidly growing parameter sizes of LLMs lead to significant deployment costs~\cite{touvron2023llama,dettmers2022gpt3int8}.
Modern LLMs often contain tens to hundreds of billions of parameters, with some mixture-of-experts models reaching the trillion-parameter scale, requiring substantial memory for inference~\cite{chowdhery2023palm,fedus2022switch,du2022glam}. During autoregressive decoding, model weights dominate memory consumption, making memory capacity a major bottleneck for inference.

Quantization addresses this challenge by approximating full-precision weights or activations with values from a discrete quantization grid, thereby reducing both storage and memory access costs~\cite{jacob2018quantization,choi2018pact}. As a result, quantization has become a key technique for improving deployment efficiency without modifying the model architecture~\cite{lin2024awq,xiao2023smoothquant}. Existing quantization methods can be broadly categorized into two paradigms: quantization-aware training (QAT) and post-training quantization (PTQ)~\cite{gholami2022survey}. QAT simulates low-precision arithmetic during training with fake-quantization operations and back-propagates through the non-differentiable quantizer via the straight-through estimator~\cite{jacob2018quantization,hubara2016binarized,bengio2013estimating}. While QAT often achieves the strongest low-bit accuracy, it requires access to training data and repeated forward--backward optimization, making it prohibitively expensive at LLM scale~\cite{dettmers2023qlora,liu2024llmqat,chen2025efficientqat,ke2024dl,malinovskii2024pvtuning}. 

In contrast, PTQ directly quantizes pretrained full-precision models using only a small calibration set and no additional training, making it a practical alternative for large-scale deployment~\cite{gong2025survey,lin2024awq,frantar2022gptq,li2025gptaq,hubara2021accurate}. Existing PTQ methods mainly improve quantization through either reshaping the weight or activation distributions, or minimizing reconstruction error on a calibration dataset. Distribution-reshaping methods, such as SmoothQuant~\cite{xiao2023smoothquant}, QuaRot~\cite{ashkboos2024quarot}, and SpinQuant~\cite{liu2024spinquant}, apply scaling or rotation transformations to make weight or activation distributions more compatible with the quantization grid. AWQ~\cite{lin2024awq} instead uses activation-aware scaling to protect salient weights from large quantization errors. Reconstruction-based methods, such as GPTQ~\cite{frantar2022gptq} and GPTAQ~\cite{li2025gptaq}, build on OBQ-style second-order approximations and perform greedy column-wise quantization with error compensation.

Existing PTQ methods differ in design, yet they typically treat the grid assignments obtained after quantization as final, as shown in Figure~\ref{fig:ptq-vs-requant}. For reconstruction-based methods such as GPTQ~\cite{frantar2022gptq} and GPTAQ~\cite{li2025gptaq}, once a column is mapped to the quantization grid, its discrete assignment is fixed, and subsequent steps compensate for the induced error through the remaining columns while leaving earlier decisions unchanged. From a discrete optimization perspective, the quantized model produced by such methods is a feasible assignment on the fixed quantization grid and remains open to further improvement after quantization.

This observation motivates us to revisit a complementary stage \emph{within} PTQ: after an upstream method has already produced an executable quantized model, its discrete assignments can still be improved while preserving the deployment format. We propose ReQuant, a backpropagation-free fixed-grid refinement procedure for this stage. ReQuant is a post-processing stage inside PTQ that takes a completed PTQ output as input, freezes bit-width, scales, zero-points, grid layout, and inference kernels, and directly revisits integer codes on that fixed grid. Through iterative refinement, ReQuant improves quantized weights from heterogeneous PTQ initializers while preserving the deployed representation.

%===============================================================================

Our main contributions are summarized as follows:
\begin{itemize}
  \item \textbf{Method.} ReQuant formulates post-quantization discrete refinement as a composable stage inside PTQ. It treats an existing PTQ solution as a feasible initialization and optimizes its integer weight codes on the fixed quantization grid, while preserving the bit-width, scale, zero-point, grid layout, and deployment format.

  \item \textbf{Analysis.} We show that ReQuant reduces the overall reconstruction loss and terminates after a finite number of accepted updates, with per-sweep complexity comparable to a single GPTAQ pass.
  
  \item \textbf{Empirical results.} Across Llama-3 8B/70B and Qwen3-14B under W4A16, W4A4, W3A4, and W2A4, ReQuant improves perplexity, KL divergence, and zero-shot accuracy over four PTQ baselines across the reported settings, with especially large gains for simple initializers and lower bit-widths; the offline budget is controllable through the number of sweeps $T$. Matched W4A16 pipeline comparisons with FlexRound~\cite{lee2023flexround} and a paired GPTQ$\pm$ReQuant study further isolate refinement gains under a backpropagation-free procedure, and a Qwen3-235B MoE experiment demonstrates large-scale feasibility.
\end{itemize}

\section{Related Work}

\paragraph{Post-training quantization.}
Post-training quantization (PTQ) maps pre-trained weights, and optionally activations, to a discrete grid using only a small calibration set, without retraining~\cite{banner2019post,shao2023omniquant,xiao2023smoothquant}. Existing PTQ methods mainly improve quantization by either reshaping weight or activation distributions to better align with the quantization grid, or minimizing reconstruction error to preserve the behavior of the full-precision model.

\emph{Distribution-reshaping} methods make weights or activations more amenable to low-bit quantization by modifying their numerical distributions before quantization. SmoothQuant~\cite{xiao2023smoothquant} uses channel-wise rescaling to migrate activation outliers into weights, thereby reducing activation quantization difficulty. OmniQuant~\cite{shao2023omniquant} learns lightweight affine transformations to reshape weight and activation distributions during calibration. QuaRot~\cite{ashkboos2024quarot} and SpinQuant~\cite{liu2024spinquant} apply orthogonal rotations to redistribute outliers across dimensions, producing representations that are more robust to low-bit quantization.

\emph{Optimization-driven} methods formulate PTQ as a reconstruction-error minimization problem and solve it through local or greedy optimization procedures.
Building on OBQ~\cite{frantar2022optimal}, GPTQ~\cite{frantar2022gptq} performs sequential column-wise quantization under a second-order approximation.
Once a column is mapped to the quantization grid, its assignment is fixed, while the resulting quantization error is compensated by closed-form updates to the remaining unquantized columns.
GPTAQ~\cite{li2025gptaq} further accounts for the activation mismatch introduced by preceding quantized layers, using the activations actually produced by the quantized model to more accurately evaluate the reconstruction error. In parallel, AWQ~\cite{lin2024awq} takes a complementary activation-aware perspective. It identifies salient weight channels using activation statistics and applies channel-wise rescaling to better preserve their contribution after quantization.

\paragraph{Construction-time continuous relaxation vs.\ post-deployment fixed-grid refinement.}
A related but distinct line of work optimizes quantization decisions while \emph{constructing} the quantized model from full-precision weights.
AdaRound~\cite{nagel2020up} learns continuous rounding variables through gradient-based optimization; BRECQ~\cite{li2021brecq} extends reconstruction-based PTQ to block-level settings; and FlexRound~\cite{lee2023flexround} jointly learns element-wise division factors and a quantization-grid scale via STE-based backpropagation for Transformers.
AdaQuant~\cite{hubara2021accurate} similarly optimizes layer-level reconstruction with continuous parameters.
These methods introduce continuous surrogate variables, optimize a relaxed objective, and then discretize to obtain the final codes.
They share a reconstruction objective with ReQuant, but operate at a different pipeline stage and optimize different variables.

ReQuant instead starts \emph{after} an upstream PTQ method has already produced an executable quantized model.
It freezes the inherited bit-width, scales, zero-points, quantization grid, storage format, and inference kernels, and directly updates integer codes on that fixed grid with exact discrete loss-change evaluations, remaining free of model backpropagation, STE, and optimizer states for learnable quantizer parameters.
Consequently, every intermediate solution remains deployable under the original format.
We therefore position ReQuant as a complementary, initializer-agnostic refinement stage inside PTQ that can be composed with AdaRound/BRECQ/FlexRound-style construction-time optimization as well as with standard heuristic or greedy PTQ pipelines.

\section{Method}

\begin{figure}[t]
  \centering
  \includegraphics[width=\linewidth]{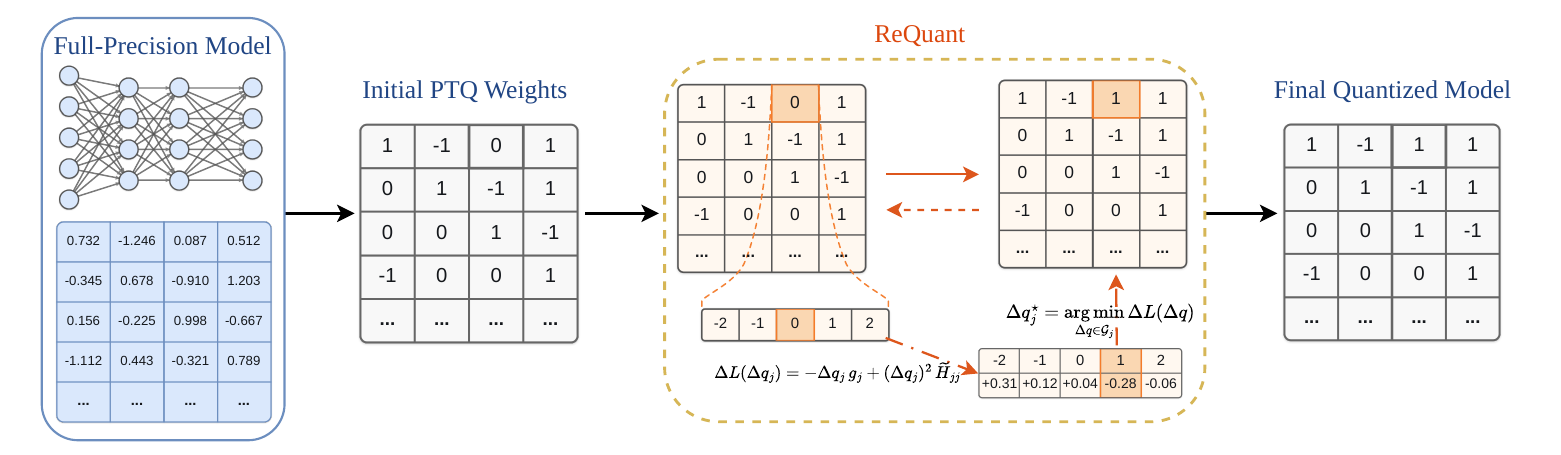}
  \caption{\textbf{Overview of ReQuant.} ReQuant refines an initial PTQ model by iteratively updating discrete weight assignments on the fixed quantization grid, improving reconstruction error while preserving the quantized format.}
  \label{fig:ptq-vs-requant}
\end{figure}

\subsection{Preliminaries}
\label{sec:prelim}

For a linear layer with full-precision weights 
$\mathbf{W}\in\mathbb{R}^{d_{\mathrm{row}}\times d_{\mathrm{col}}}$ 
and calibration activations 
$\mathbf{X}\in\mathbb{R}^{d_{\mathrm{col}}\times m}$, 
classical layer-wise PTQ is commonly formulated as minimizing the reconstruction error
\begin{equation}
  \min\; 
  \|\mathbf{W}\mathbf{X}-\mathbf{W}_q\mathbf{X}\|_F^2.
  \label{eq:ptq-objective}
\end{equation}

However, at inference time, the input to a given layer is produced by preceding quantized layers rather than by the full-precision model. 
To model this activation mismatch, GPTAQ~\cite{li2025gptaq} replaces the full-precision activations $\mathbf{X}$ in the quantized branch with the activations $\widetilde{\mathbf{X}}$ observed under the quantized prefix, yielding the activation-aware objective
\begin{equation}
  \min\; 
  \mathcal{L}(\mathbf{W}_q)
  \;=\;
  \|\mathbf{W}\mathbf{X}-\mathbf{W}_q\widetilde{\mathbf{X}}\|_F^2 .
  \label{eq:gptaq-objective}
\end{equation}
Both $\mathbf{X}$ and $\widetilde{\mathbf{X}}$ can be collected by forward passes on the calibration set, using the full-precision and partially quantized models, respectively.
% Both Eq.\ref{eq:ptq-objective} and Eq.\ref{eq:gptaq-objective} define discrete optimization problems over the quantization grid $\mathcal G$. Since each entry of $\mathbf{W}q$ must take a value from $\mathcal G$, the feasible space contains $|\mathcal G|^{d{\mathrm{row}} d_{\mathrm{col}}}$ possible assignments. This combinatorial search space grows exponentially with the number of weights, making exhaustive search infeasible for modern LLM layers. Moreover, the fixed-grid constraint prevents direct application of standard gradient-based optimization in continuous space.

\paragraph{Notation.}
For a matrix $\mathbf{W}\in\mathbb{R}^{d_{\mathrm{row}}\times d_{\mathrm{col}}}$, the first subscript indexes rows and the second indexes columns: $\mathbf{W}_{i,:}$ denotes its $i$-th row and $\mathbf{W}_{:,j}$ denotes its $j$-th column. For an ordered index set $R$, $\mathbf{W}_{i,R}$ denotes the entries of row $i$ restricted to the columns in $R$, and $\mathbf{W}_{R,R}$ denotes the corresponding principal submatrix when the matrix is square. The same convention applies to $\widetilde{\mathbf{H}}:=\widetilde{\mathbf{X}}\widetilde{\mathbf{X}}^\top$ and other matrices appearing below.

\subsection{ReQuant Method}
\label{sec:ReQuant_Method}
The framework of ReQuant is shown in Figure~\ref{fig:ptq-vs-requant}. The core idea is that existing PTQ methods typically stop after a single quantization procedure and treat the resulting on-grid assignment as final, whereas ReQuant uses this assignment as initialization and continues to optimize the discrete weight values on the same quantization grid.

\paragraph{Optimization objective.}
After the quantization parameters (e.g., scale and zero-point) are fixed, we consider the optimization problem on the induced discrete grid. Our objective is to further optimize the quantized weights:
\begin{equation}
  \min_{\mathbf{W}_q \in \mathcal G}
  \mathcal{L}(\mathbf{W}_q)
  \;=\; \|\mathbf{W}\mathbf{X}-\mathbf{W}_q\widetilde{\mathbf{X}}\|_F^2 .
  \label{eq:discrete_obj}
\end{equation}
where $\mathcal G$ is the fixed quantization grid. We refer to $\mathcal{L}(\mathbf{W}_q)$ as the layer-wise reconstruction objective. In this work, we adopt Eq.~\ref{eq:discrete_obj} as the refinement objective and optimize the quantized weights directly over the discrete grid.

\paragraph{Row-wise decomposition.}
\label{sec:row-decomposition}
The key observation is that each output row of a linear layer is produced by the
corresponding weight row independently. In particular, the $i$-th row of
$\mathbf{W}_q$ only affects the $i$-th row of the quantized output
$\mathbf{W}_q\widetilde{\mathbf X}$. Therefore, the reconstruction loss decomposes as
\begin{equation}
  \|\mathbf{W}\mathbf{X}-\mathbf{W}_q\widetilde{\mathbf X}\|_F^2
  =
  \sum_{i=1}^{d_{\mathrm{row}}}
  \|\mathbf{W}_{i,:}\mathbf{X}
  -\mathbf{W}_{q,i,:}\widetilde{\mathbf X}\|_2^2 .
  \label{eq:row-decomposition}
\end{equation}

Updating a single coordinate $W_{q,i,j}$ affects only the $i$-th term in Eq.~\ref{eq:row-decomposition}; the losses of all other rows remain unchanged. Thus, reducing $L$ for any single row directly decreases the layer-level loss $\mathcal{L}(\mathbf{W}_q)$, and ReQuant can refine each row independently and in parallel. In the rest of this section we derive the update rule for a single row and omit the row index for clarity.

\paragraph{Row-level reconstruction loss.}
Following the row-wise decomposition above, we focus on a single row and omit the row index. Let $\mathbf{w}=\mathbf{W}_{i,:}$ denote the full-precision weight row, $\mathbf{q}=\mathbf{W}_{q,i,:}$ denote its current quantized counterpart, and $\mathbf{e}=\mathbf{w}-\mathbf{q}$ denote the corresponding quantization error. By Eq.~\ref{eq:row-decomposition}, the contribution of this row to $\mathcal{L}(\mathbf{W}_q)$ is
\begin{equation}
  L(\mathbf{e})
  =
  \|\mathbf{w}\mathbf{X}-\mathbf{q}\widetilde{\mathbf{X}}\|_2^2
  =
  \|\mathbf{e}\widetilde{\mathbf{X}}-\mathbf{w}\Delta\mathbf{X}\|_2^2 ,
  \label{eq:row_objective}
\end{equation}
where $\Delta\mathbf{X}=\widetilde{\mathbf X}-\mathbf{X}$ denotes the activation mismatch induced by earlier quantized layers. Eq.~\ref{eq:row_objective} rewrites the row-level loss as a function of the quantization error $\mathbf{e}$, which is the form on which ReQuant operates.

We define the gradient vector
$\mathbf{g}$ as the gradient of $L$ with respect to $\mathbf{e}$:
\begin{equation}
  \mathbf{g}
  =
  \nabla_{\mathbf e} L(\mathbf{e})
  =
  2(\mathbf{e}\widetilde{\mathbf H}-\mathbf{w}\mathbf{B}),
  \label{eq:row-score}
\end{equation}
where $\mathbf{B}=\Delta\mathbf{X}\widetilde{\mathbf X}^\top$ and $\widetilde{\mathbf H}=\widetilde{\mathbf X}\widetilde{\mathbf X}^\top\succeq 0$ is the row-level Hessian of $L$ with respect to $\mathbf{e}$. Consequently, $L(\mathbf{e})$ is a convex quadratic function of $\mathbf{e}$, and the curvature along coordinate $j$ is $\widetilde H_{jj}$. If the
quantized value at coordinate $j$ is moved by an on-grid step $\Delta q_j$, then
the quantization error becomes $\mathbf{e}'=\mathbf{e}-\Delta q_j\mathbf{u}_j$, and the loss change is
\begin{equation}
  \Delta L(\Delta q_j)
  =
  L(\mathbf{e}')-L(\mathbf{e})
  =
  -\Delta q_j\, g_j
  +(\Delta q_j)^2\,\widetilde H_{jj}.
  \label{eq:coordinate_delta}
\end{equation}
The derivation is given in Appendix~\ref{app:requant-coordinate-update}. This computation reduces the cost of evaluating the reconstruction error. Once the current row loss $L$ and vector $\mathbf{g}$ are computed, the effect of any subsequent coordinate update can be obtained directly from Eq.~\ref{eq:row-score} and Eq.~\ref{eq:coordinate_delta}. Therefore, during refinement, ReQuant only needs to maintain the current row loss $L$ and vector $\mathbf{g}$, which substantially reduces the cost of evaluating candidate updates.

\paragraph{Coordinate update.}
\label{sec:coord-update}

Let $s_j$, $o_j$, and $z_j$ denote the scale, zero-point, and quantized integer
code of coordinate $j$, so that $q_j=(z_j-o_j)s_j$. Starting from the current integer
code $z_j$, ReQuant considers integer offsets $k$ such that the updated code
$z_j+k$ remains within the representable range. Therefore, any candidate update
has the form $\Delta q_j=k s_j$, and the feasible set at coordinate $j$ is
\begin{equation}
  \mathcal G_j=  \{\, k s_j : k\in\mathbb Z,\; k\ne 0,\; z_{\min}\le z_j+k\le z_{\max} \,\}.
  \label{eq:feasible_steps}
\end{equation}
In practice we further restrict the search to a local neighborhood of size $K$,
\[
  \mathcal G_j^{(K)}=\{\,ks_j\in\mathcal G_j: 1\le |k|\le K\,\},
\]
and score candidates with Eq.~\ref{eq:coordinate_delta}, selecting
\begin{equation}
  \Delta q_j^\star = \arg\min_{\Delta q\in\mathcal G_j^{(K)}}
  \Delta L(\Delta q).
  \label{eq:best_move}
\end{equation}
This step can be viewed as a discrete coordinate descent update. With all other
coordinates fixed, ReQuant moves coordinate $j$ to the best feasible grid point
in the $K$-neighborhood.

ReQuant accepts a move whenever it strictly decreases the row loss, i.e.,
$\Delta L(\Delta q_j^\star)<0$. Upon acceptance, ReQuant updates coordinate $j$
and increments the row loss by $L\leftarrow L+\Delta L(\Delta q_j^\star)$. The
gradient vector is then refreshed incrementally as
\begin{equation}
  \mathbf g
  \leftarrow
  \mathbf g-2\Delta q_j^\star\widetilde{\mathbf H}_{j,:}.
  \label{eq:score_update}
\end{equation}
This follows from $\mathbf g=2(\mathbf e\widetilde{\mathbf H}-\mathbf w\mathbf B)$, since an accepted move changes $\mathbf e$ only at coordinate $j$. Therefore, $\mathbf g$ can be updated in $O(d_{\mathrm{col}})$ time using the $j$-th row of $\widetilde{\mathbf H}$.

Because each accepted update changes $\mathbf e$ and hence the score vector $\mathbf g$, a coordinate that appears locally fixed in one pass can become improvable after other coordinates have changed. Thus, a single greedy traversal over $j=1,\ldots,d_{\mathrm{col}}$ leaves additional discrete improvements available for the coupled quadratic objective $L(\mathbf e)$. ReQuant therefore repeats the coordinate cycle for a fixed number of sweeps $T$, as summarized in Algorithm~\ref{alg:refine}, providing a controllable compute--accuracy trade-off.

\begin{algorithm}[t]
  \caption{Post-Quantization Discrete Coordinate Refinement}
  \label{alg:refine}
  \small
  \begin{algorithmic}[1]
    \STATE \textbf{Input:} full-precision row $\mathbf w$, initial quantized row $\mathbf q$, grid information, $\widetilde{\mathbf H}$, $\mathbf B$, neighborhood size $K$, sweeps $T$
    \STATE $\mathbf e \leftarrow \mathbf w-\mathbf q$, \quad $\mathbf g \leftarrow 2(\mathbf e\widetilde{\mathbf H}-\mathbf w\mathbf B)$
   \STATE $L \leftarrow \|\mathbf e\widetilde{\mathbf X}-\mathbf w\Delta\mathbf X\|_2^2$
    \FOR{$t=1,\dots,T$}
      \FOR{$j=1,\dots,d_{\mathrm{col}}$}
        \STATE $\Delta q_j^\star \leftarrow \arg\min_{\Delta q_j\in\mathcal G_j^{(K)}} \Delta L(\Delta q_j)$
        \IF{$\Delta L(\Delta q_j^\star)<0$}
          \STATE $q_j \leftarrow q_j+\Delta q_j^\star$
          \STATE $\mathbf g\leftarrow \mathbf g-2\Delta q_j^\star\widetilde{\mathbf H}_{j,:}$
          \STATE $L \leftarrow L+\Delta L(\Delta q_j^\star)$
        \ENDIF
      \ENDFOR
    \ENDFOR
    \STATE \textbf{Return:} refined quantized row $\mathbf q$
  \end{algorithmic}
\end{algorithm}

\subsection{Analysis}
\paragraph{Convergence.}
\label{sec:analysis-convergence}
ReQuant performs discrete coordinate descent over a finite quantization grid, searching the $K$-neighborhood $\mathcal G_j^{(K)}$ at each coordinate. At each step, a candidate update is accepted whenever $\Delta L(\Delta q_j^\star)<0$ (Algorithm~\ref{alg:refine}). Hence, every accepted update strictly decreases the corresponding row-level loss $L$. Since the layer objective decomposes as $\mathcal{L}(\mathbf{W}_q)=\sum_i L(\mathbf{e}_i)$ and each update affects a single row, it also strictly decreases $\mathcal{L}(\mathbf{W}_q)$. The feasible set is finite, and the objective strictly decreases after each accepted update, so each accepted move yields a distinct quantized state with strictly lower reconstruction loss. If refinement continues until every $K$-neighborhood move has been exhausted, the returned solution is a coordinate-wise local optimum within that neighborhood; in practice a fixed budget of $T$ sweeps already yields strong empirical improvements while keeping offline cost controllable. A full proof of the infinite-sweep case is provided in Appendix~\ref{app:requant-termination}.

As the calibration set increases, the empirical activation statistics in $\widetilde{\mathbf H}$ and $\mathbf B$ provide increasingly accurate estimates of their population counterparts. As a result, the empirical refinement objective better approximates the population-level objective, and the solution obtained by ReQuant becomes better aligned with the population-level on-grid optimum.

\paragraph{Reconstruction loss versus downstream metrics.}
The analysis guarantees monotone decrease of the empirical layer-wise objective along accepted updates, and neighborhood-local optimality when the $K$-search is run to exhaustion. Reducing layer reconstruction error is a standard proxy in PTQ~\cite{nagel2020up,li2021brecq,frantar2022gptq}: if the remaining network is locally Lipschitz, smaller layer residuals tighten a bound on later hidden-state perturbations. Consistent with this proxy, our experiments show that reconstruction-driven refinement typically improves held-out PPL, KL, and zero-shot accuracy, which we report as empirical outcomes of Algorithm~\ref{alg:refine}.

\paragraph{Efficiency.}
ReQuant evaluates candidate updates from cached statistics rather than recomputing the full reconstruction loss each time. Using the row-wise decomposition and the closed-form coordinate score in Eq.~\ref{eq:coordinate_delta}, it scores on-grid moves efficiently and refreshes the score vector after each accepted move with a rank-one update. Since output rows are independent, they can also be refined in parallel while sharing the same activation statistics. For a layer with $T$ refinement sweeps, the dominant cost is $O(m d_{\mathrm{col}}^2 + (T+1)d_{\mathrm{row}}d_{\mathrm{col}}^2)$ under low-bit quantization, where each coordinate has a small number of feasible moves. Appendix~\ref{app:requant-complexity} provides the full complexity analysis.

\section{Experiments}

\subsection{Experimental Setup}
\label{sec:exp-setup}

\paragraph{Overview.}
We organize our experiments as follows. First, we use GPTAQ as the primary baseline and evaluate perplexity and downstream task performance on Qwen3~\cite{yang2025qwen3} and Llama-3~\cite{grattafiori2024llama3}. We then examine scale, including 8B, 70B, and a 235B MoE setting (Section~\ref{sec:qwen235b}). Finally, we evaluate generality across PTQ initializers (RTN, AWQ~\cite{lin2024awq}, GPTQ~\cite{frantar2022gptq}, GPTAQ~\cite{li2025gptaq}), cost--benefit via $T$, and comparisons with FlexRound.

\paragraph{Implementation details.}
We implement ReQuant with Hugging Face Transformers~\cite{wolf2020transformers} and PyTorch~\cite{paszke2019pytorch}. We use WikiText-2~\cite{merity2016wikitext} as the calibration dataset and sample 512 sequences of length 2048 to collect activation statistics. Unless otherwise specified, weights are quantized with per-channel asymmetric quantization, and activations are quantized with per-tensor asymmetric quantization. Following GPTAQ~\cite{li2025gptaq}, activation quantizers are calibrated and applied before weight quantization, so ReQuant optimizes weights under the activations produced by the quantized model. By default, we set the number of ReQuant refinement sweeps to $T{=}4$ and evaluate $K{=}2$ neighboring grid moves in each direction for every coordinate. Unless noted otherwise, main tables use data-parallel calibration on eight NVIDIA GeForce RTX~4090 GPUs; Llama-3~70B uses four NVIDIA B200 GPUs. FlexRound comparisons (Section~\ref{sec:flexround}) are timed on an NVIDIA RTX PRO~6000 for memory-matched fairness, and the Qwen3-235B experiment (Section~\ref{sec:qwen235b}) uses eight NVIDIA H200 GPUs.

\paragraph{Evaluation.}
We evaluate language modeling performance by reporting perplexity on WikiText-2~\cite{merity2016wikitext}, UltraChat-2k~\cite{ding2023enhancing}, and NuminaMath~\cite{li2024numinamath}. For downstream evaluation, we report accuracy on ten standard commonsense reasoning and language understanding benchmarks, including ARC-Challenge (ARC-C), ARC-Easy (ARC-E), BoolQ, CEval, HellaSwag, LAMBADA, OpenBookQA (OBQA), PIQA, SocialIQA (SIQA), and Winogrande~\cite{clark2018think,clark2019boolq,huang2023ceval,zellers2019hellaswag,paperno2016lambada,mihaylov2018can,bisk2020piqa,sap2019socialiqa,sakaguchi2021winogrande}. We use the average accuracy across these tasks as the overall downstream score.

\subsection{Main results}
\label{sec:main-results}

\begin{table}[t]
  \centering
  \caption{Zero-shot accuracy (\%) at W4A16 and W4A4; both settings use QuaRot~\cite{ashkboos2024quarot} rotation (W4A16 without QuaRot is in Appendix~\ref{app:w4a16-no-rotation}). \textbf{Avg.} is the mean over ten tasks; the better entry of each pair is bolded. Corresponding KL/PPL are reported in Table~\ref{tab:requant-w4a16-kl-ppl} (Appendix~\ref{app:kl-ppl-details}) and Table~\ref{tab:requant-kl-ppl-main} (Appendix~\ref{app:kl-ppl-w4a4-8b-14b}).}
  \label{tab:requant-acc-main}
  \scriptsize
  \setlength{\tabcolsep}{2pt}
  \resizebox{\textwidth}{!}{%
  \begin{tabular}{@{}c|l|cccccccccc|c@{}}
    \toprule
    \textbf{Precision} & \textbf{Method} & \textbf{Arc C} & \textbf{Arc E} & \textbf{BoolQ} & \textbf{CEval} & \textbf{HellaSwag} & \textbf{LAMBADA} & \textbf{OBQA} & \textbf{PIQA} & \textbf{SIQA} & \textbf{Winogrande} & \textbf{Avg.$\uparrow$} \\
    \midrule
    \multicolumn{13}{l}{\emph{Llama-3 8B}} \\
    \cmidrule(lr){1-13}
    FP                          & --                & 53.58          & 77.69          & 81.25          & 48.44          & 79.17          & 75.70          & 45.00          & 80.90          & 46.88          & 73.24          & 66.19 \\
    \cmidrule(lr){1-13}
    \multirow{8}{*}{W4A16}      & RTN               & 47.53          & 69.74          & 78.01          & 36.63          & 75.78          & 73.78          & 43.00          & 77.91          & 45.04          & 70.40          & 61.78 \\
                                & RTN + ReQuant     & \textbf{50.68} & \textbf{75.93} & \textbf{80.18} & \textbf{41.98} & \textbf{77.06} & \textbf{73.86} & \textbf{43.20} & \textbf{79.16} & \textbf{45.39} & \textbf{72.93} & \textbf{64.04} \\
    \cmidrule(lr){2-13}
                                & AWQ               & 50.34          & \textbf{77.53} & 79.05          & 38.86          & 77.34          & \textbf{74.36} & \textbf{44.80} & \textbf{80.09} & 45.75          & 71.82          & 63.99 \\
                                & AWQ + ReQuant     & \textbf{51.79} & 75.59          & \textbf{82.11} & \textbf{44.65} & \textbf{77.72} & 73.92          & 44.20          & 79.43          & \textbf{46.42} & \textbf{72.30} & \textbf{64.81} \\
    \cmidrule(lr){2-13}
                                & GPTQ              & 51.88          & 77.65          & 80.76          & 45.32          & \textbf{78.13} & 74.99          & \textbf{44.60} & 79.49          & 45.80          & \textbf{73.56} & 65.22 \\
                                & GPTQ + ReQuant    & \textbf{54.18} & \textbf{77.69} & \textbf{82.14} & \textbf{45.47} & 78.01          & \textbf{75.12} & \textbf{44.60} & \textbf{79.98} & \textbf{46.06} & 72.22          & \textbf{65.55} \\
    \cmidrule(lr){2-13}
                                & GPTAQ             & \textbf{53.33} & 77.95          & \textbf{81.80} & 44.73          & 77.97          & 75.28          & 44.40          & 80.20          & 45.85          & \textbf{73.48} & 65.50 \\
                                & GPTAQ + ReQuant   & \textbf{53.33} & \textbf{78.16} & 80.86          & \textbf{47.47} & \textbf{78.17} & \textbf{75.49} & \textbf{45.20} & \textbf{80.36} & \textbf{46.21} & 72.30          & \textbf{65.76} \\
    \cmidrule(lr){1-13}
    \multirow{8}{*}{W4A4}       & RTN               & 39.85          & 62.58          & 67.40          & 31.05          & 66.16          & 51.31          & 37.80          & 70.51          & 42.37          & 64.25          & 53.33 \\
                                & RTN + ReQuant     & \textbf{46.25} & \textbf{74.58} & \textbf{78.07} & \textbf{38.11} & \textbf{74.46} & \textbf{70.02} & \textbf{44.00} & \textbf{77.58} & \textbf{45.65} & \textbf{70.72} & \textbf{61.94} \\
    \cmidrule(lr){2-13}
                                & AWQ               & 37.97          & 60.44          & 65.93          & 28.97          & 68.89          & 60.59          & 37.80          & 71.11          & 41.50          & 64.01          & 53.72 \\
                                & AWQ + ReQuant     & \textbf{44.28} & \textbf{69.95} & \textbf{73.67} & \textbf{34.32} & \textbf{73.22} & \textbf{66.93} & \textbf{42.40} & \textbf{75.90} & \textbf{44.73} & \textbf{67.32} & \textbf{59.27} \\
    \cmidrule(lr){2-13}
                                & GPTQ              & 46.93          & 74.24          & 75.81          & 36.92          & 74.59          & 70.13          & \textbf{42.60} & 76.77          & 45.29          & \textbf{69.61} & 61.29 \\
                                & GPTQ + ReQuant    & \textbf{47.35} & \textbf{75.00} & \textbf{76.67} & \textbf{37.67} & \textbf{75.46} & \textbf{71.10} & 42.00          & \textbf{78.07} & \textbf{45.60} & 68.43          & \textbf{61.82} \\
    \cmidrule(lr){2-13}
                                & GPTAQ             & 47.27          & \textbf{75.88} & \textbf{78.62} & 37.59          & 75.57          & 70.50          & 42.00          & 78.07          & 45.70          & 68.51          & 61.97 \\
                                & GPTAQ + ReQuant   & \textbf{50.09} & 74.16          & 77.58          & \textbf{39.38} & \textbf{75.69} & \textbf{70.99} & 42.00          & \textbf{78.51} & \textbf{45.75} & \textbf{69.61} & \textbf{62.38} \\
    \midrule
    \multicolumn{13}{l}{\emph{Qwen3-14B}} \\
    \cmidrule(lr){1-13}
    FP                          & --                & 60.32          & 82.95          & 89.30          & 82.32          & 78.82          & 67.82          & 46.40          & 79.87          & 52.05          & 72.53          & 71.24 \\
    \cmidrule(lr){1-13}
    \multirow{8}{*}{W4A16}      & RTN               & 55.20          & 78.32          & 86.45          & 78.38          & 75.82          & 67.09          & 42.20          & 79.11          & 48.00          & 69.69          & 68.03 \\
                                & RTN + ReQuant     & \textbf{59.90} & \textbf{82.45} & \textbf{88.47} & \textbf{79.20} & \textbf{77.72} & \textbf{67.75} & \textbf{46.60} & \textbf{80.03} & \textbf{50.61} & \textbf{72.69} & \textbf{70.54} \\
    \cmidrule(lr){2-13}
                                & AWQ               & 57.76          & 81.86          & 88.07          & 78.16          & 76.17          & 62.72          & \textbf{45.80} & 79.33          & 50.56          & 71.27          & 69.17 \\
                                & AWQ + ReQuant     & \textbf{59.30} & \textbf{82.07} & \textbf{89.60} & \textbf{80.46} & \textbf{77.74} & \textbf{66.25} & 45.40          & \textbf{79.71} & \textbf{51.94} & \textbf{72.14} & \textbf{70.46} \\
    \cmidrule(lr){2-13}
                                & GPTQ              & \textbf{59.73} & \textbf{82.03} & 88.04          & 79.94          & \textbf{77.97} & 68.29          & 45.00          & 79.43          & 51.69          & \textbf{73.09} & 70.52 \\
                                & GPTQ + ReQuant    & 59.47          & 81.40          & \textbf{88.96} & \textbf{81.13} & 77.92          & \textbf{68.35} & \textbf{46.00} & \textbf{79.82} & \textbf{51.79} & 72.06          & \textbf{70.69} \\
    \cmidrule(lr){2-13}
                                & GPTAQ             & 58.96          & \textbf{82.32} & \textbf{89.02} & \textbf{80.31} & 77.45          & 67.79          & \textbf{45.40} & \textbf{80.36} & 51.28          & \textbf{71.98} & 70.49 \\
                                & GPTAQ + ReQuant   & \textbf{60.07} & 82.11          & 88.47          & 80.01          & \textbf{77.76} & \textbf{68.50} & 44.80          & 79.87          & \textbf{51.74} & \textbf{71.98} & \textbf{70.53} \\
    \cmidrule(lr){1-13}
    \multirow{8}{*}{W4A4}       & RTN               & 48.72          & 73.40          & 84.98          & 74.15          & 72.26          & 64.20          & 41.40          & 76.55          & 46.21          & 67.80          & 64.97 \\
                                & RTN + ReQuant     & \textbf{54.95} & \textbf{79.92} & \textbf{86.79} & \textbf{74.67} & \textbf{74.45} & \textbf{65.46} & \textbf{43.20} & \textbf{77.58} & \textbf{49.85} & \textbf{69.22} & \textbf{67.61} \\
    \cmidrule(lr){2-13}
                                & AWQ               & 51.37          & 74.20          & 85.63          & 73.55          & 73.26          & 62.22          & 42.40          & 77.09          & 47.65          & 66.14          & 65.35 \\
                                & AWQ + ReQuant     & \textbf{57.85} & \textbf{80.26} & \textbf{87.06} & \textbf{76.30} & \textbf{74.10} & \textbf{63.50} & \textbf{43.60} & \textbf{77.48} & \textbf{48.72} & \textbf{67.56} & \textbf{67.64} \\
    \cmidrule(lr){2-13}
                                & GPTQ              & 56.83          & 77.48          & \textbf{86.70} & 76.37          & \textbf{74.69} & 65.38          & \textbf{44.80} & 78.18          & \textbf{48.77} & \textbf{70.32} & 67.95 \\
                                & GPTQ + ReQuant    & \textbf{57.25} & \textbf{79.92} & 86.39          & \textbf{77.04} & 74.57          & \textbf{65.79} & 43.80          & \textbf{78.67} & 48.21          & 68.67          & \textbf{68.03} \\
    \cmidrule(lr){2-13}
                                & GPTAQ             & 55.20          & 78.41          & 86.73          & 74.00          & \textbf{74.25} & 64.23          & \textbf{42.40} & 77.15          & \textbf{48.82} & \textbf{68.98} & 67.02 \\
                                & GPTAQ + ReQuant   & \textbf{55.89} & \textbf{79.63} & \textbf{87.77} & \textbf{76.45} & 74.08          & \textbf{65.92} & 41.00          & \textbf{78.29} & 48.16          & 68.11          & \textbf{67.53} \\
    \bottomrule
  \end{tabular}}
\end{table}

\paragraph{Results on Llama-3 8B and Qwen3-14B.}
We evaluate ReQuant on Llama-3 8B and Qwen3-14B under W4A16 and W4A4 quantization. We use four representative PTQ initializers: RTN~\cite{banner2019post}, AWQ~\cite{lin2024awq}, GPTQ~\cite{frantar2022gptq}, and GPTAQ~\cite{li2025gptaq}, covering both heuristic and greedy reconstruction-based quantization methods. In Table~\ref{tab:requant-acc-main}, both W4A16 and W4A4 use QuaRot~\cite{ashkboos2024quarot} rotations; a W4A16 ablation without QuaRot is reported in Appendix~\ref{app:w4a16-no-rotation}. ReQuant is applied only after the initial PTQ stage: the quantization grid, scale, zero-point, and bit-width remain fixed, and only the integer code at each coordinate is updated. Table~\ref{tab:requant-acc-main} reports the average zero-shot accuracy over ten downstream tasks across all model, precision, and initializer combinations. We defer the corresponding KL divergence and perplexity (PPL) results on WikiText-2, UltraChat-2k, and NuminaMath to Appendix~\ref{app:kl-ppl-details} and Appendix~\ref{app:kl-ppl-w4a4-8b-14b}, report runtime measurements in Appendix~\ref{app:runtime-qwen}, and analyze how WikiText-2 PPL, KL, and average zero-shot accuracy vary with the calibration sample count for Llama-3 8B at W4A4 in Appendix~\ref{app:gptaq-calibration-samples}.

Table~\ref{tab:requant-acc-main} shows that ReQuant improves average downstream accuracy across both precision settings, both model families, and all four PTQ initializers in the reported pairs. The gains are larger for simpler initializers such as RTN and AWQ, indicating that many errors left by basic quantizers can still be corrected on the fixed grid. For example, under W4A16 on Qwen3-14B, ReQuant improves RTN by $+2.51$ average accuracy points. Under the more challenging W4A4 setting, where both weights and activations are quantized to 4 bits, the improvement becomes more pronounced: on Llama-3 8B, ReQuant improves RTN by $+8.61$ average accuracy points.

ReQuant also improves stronger initializers such as GPTQ and GPTAQ, confirming that residual fixed-grid assignment error remains exploitable even after advanced PTQ construction. The offline budget remains controllable through $T$, so practitioners can allocate more sweeps when larger gains are desired.

Notably, ReQuant substantially narrows the gap between simple and advanced PTQ methods. Under W4A4, RTN refined by ReQuant nearly matches GPTAQ on Llama-3 8B in average accuracy, and surpasses GPTAQ on Qwen3-14B. This suggests that a substantial share of the performance gap among PTQ methods comes from correctable discrete assignment errors that ReQuant can recover on the fixed grid.

\begin{table}[t]
  \centering
  \caption{Zero-shot accuracy (\%) on Llama-3 70B at W4A4 with QuaRot~\cite{ashkboos2024quarot} rotation. \textbf{Avg.} is the mean over ten tasks; the better entry of each pair is bolded. Corresponding KL/PPL are reported in Table~\ref{tab:requant-70b-kl-ppl} of Appendix~\ref{app:kl-ppl-70b}.}
  \label{tab:requant-70b-acc}
  \footnotesize
  \setlength{\tabcolsep}{2.5pt}
  \resizebox{\textwidth}{!}{%
  \begin{tabular}{@{}l|cccccccccc|c@{}}
    \toprule
    \textbf{Method}     & \textbf{Arc C} & \textbf{Arc E} & \textbf{BoolQ} & \textbf{CEval} & \textbf{HellaSwag} & \textbf{LAMBADA} & \textbf{OBQA} & \textbf{PIQA} & \textbf{SIQA} & \textbf{Winogrande} & \textbf{Avg.$\uparrow$} \\
    \midrule
    FP                  & 64.25          & 85.94          & 85.38          & 64.64          & 84.96          & 79.33          & 49.20          & 84.39          & 50.61          & 80.51          & 72.92 \\
    \cmidrule(lr){1-12}
    RTN                 & 22.18          & 33.63          & 43.76          & 23.11          & 31.10          & 7.96           & 25.40          & 55.44          & 35.47          & 49.72          & 32.78 \\
    RTN + ReQuant       & \textbf{55.12} & \textbf{80.13} & \textbf{84.50} & \textbf{46.43} & \textbf{79.83} & \textbf{77.97} & \textbf{44.40} & \textbf{80.96} & \textbf{46.16} & \textbf{73.80} & \textbf{66.93} \\
    \cmidrule(lr){1-12}
    AWQ                 & 34.04          & 57.41          & 64.37          & 23.11          & 57.46          & 36.31          & 35.60          & 68.82          & 39.25          & 58.96          & 47.53 \\
    AWQ + ReQuant       & \textbf{54.35} & \textbf{80.81} & \textbf{83.36} & \textbf{44.87} & \textbf{78.13} & \textbf{75.94} & \textbf{45.20} & \textbf{80.69} & \textbf{45.65} & \textbf{74.90} & \textbf{66.39} \\
    \cmidrule(lr){1-12}
    GPTQ                & 43.77          & 69.02          & 72.63          & 30.91          & 71.68          & 63.05          & 42.00          & 75.84          & 41.40          & 70.56          & 58.09 \\
    GPTQ + ReQuant      & \textbf{54.18} & \textbf{79.29} & \textbf{82.84} & \textbf{47.10} & \textbf{77.74} & \textbf{76.25} & \textbf{44.00} & \textbf{79.71} & \textbf{46.42} & \textbf{73.16} & \textbf{66.07} \\
    \cmidrule(lr){1-12}
    GPTAQ               & \textbf{55.72} & 79.84          & 81.96          & 46.51          & 78.79          & 73.74          & 43.80          & \textbf{80.85} & 45.45          & \textbf{75.37} & 66.20 \\
    GPTAQ + ReQuant     & 55.03          & \textbf{80.43} & \textbf{84.40} & \textbf{47.77} & \textbf{79.15} & \textbf{76.77} & \textbf{44.20} & \textbf{80.85} & \textbf{46.16} & 75.30          & \textbf{67.01} \\
    \bottomrule
  \end{tabular}}
\end{table}

\paragraph{Results on Llama-3 70B at W4A4.} We further evaluate ReQuant on Llama-3 70B under the challenging W4A4 setting with QuaRot~\cite{ashkboos2024quarot} rotation. ReQuant is applied to the same four PTQ initializers: RTN, AWQ, GPTQ, and GPTAQ. Table~\ref{tab:requant-70b-acc} reports zero-shot accuracy across ten downstream tasks, and the corresponding KL divergence and perplexity results are provided in Table~\ref{tab:requant-70b-kl-ppl} of Appendix~\ref{app:kl-ppl-70b}.  
 
At the 70B scale, ReQuant improves all four initializers on average accuracy. On GPTAQ, average accuracy rises from $66.20\%$ to $67.01\%$ ($+0.81$), with lower perplexity and KL as well, showing that even a strong activation-aware initializer retains correctable fixed-grid error. Starting from RTN, ReQuant reaches $66.93\%$ average accuracy---close to GPTAQ+ReQuant---and obtains better KL/PPL than GPTAQ+ReQuant on the reported evaluation sets. Thus fixed-grid refinement can lift a simple initializer to a competitive final discrete assignment.

\subsection{Lower-bit weight quantization}
\label{sec:low-bit}

We further evaluate ReQuant under more aggressive settings, W3A4 and W2A4, on Llama-3 8B with QuaRot~\cite{ashkboos2024quarot} rotation. We focus on GPTQ and GPTAQ, the strongest reconstruction-based initializers at these bit-widths. Table~\ref{tab:requant-low-bit} reports zero-shot accuracy; the corresponding KL/PPL results are deferred to Appendix~\ref{app:low-bit-kl-ppl}. ReQuant improves both baselines on average accuracy, with larger gains at lower bit-width. At W3A4, ReQuant improves GPTQ from $57.10$ to $58.61$ ($+1.51$) and further improves GPTAQ from $58.65$ to $58.71$. At W2A4, the gains are much larger: GPTQ improves from $35.88$ to $41.00$ ($+5.12$), matching GPTAQ ($40.90$), while GPTAQ itself rises to $41.91$. Across both settings, ReQuant narrows the GPTQ--GPTAQ gap and nearly closes it at W2A4, indicating that a large share of the very-low-bit gap is recoverable discrete assignment error on the fixed grid.

\begin{table}[t]
  \centering
  \caption{Zero-shot accuracy (\%) of ReQuant on Llama-3 8B at W3A4 and W2A4 with QuaRot~\cite{ashkboos2024quarot} rotation. \textbf{Avg.} is the mean over ten tasks; the better entry of each pair is bolded. Corresponding KL/PPL are in Table~\ref{tab:requant-low-bit-kl-ppl} (Appendix~\ref{app:low-bit-kl-ppl}).}
  \label{tab:requant-low-bit}
  \footnotesize
  \setlength{\tabcolsep}{2.5pt}
  \resizebox{\textwidth}{!}{%
  \begin{tabular}{@{}c|l|cccccccccc|c@{}}
    \toprule
    \textbf{Precision} & \textbf{Method} & \textbf{Arc C} & \textbf{Arc E} & \textbf{BoolQ} & \textbf{CEval} & \textbf{HellaSwag} & \textbf{LAMBADA} & \textbf{OBQA} & \textbf{PIQA} & \textbf{SIQA} & \textbf{Winogrande} & \textbf{Avg.$\uparrow$} \\
    \midrule
    \multirow{4}{*}{W3A4}
                            & GPTQ             & 40.02          & 65.82          & 74.53          & \textbf{31.43} & \textbf{70.08} & 64.53          & 38.60          & \textbf{74.92} & \textbf{44.11} & 66.93          & 57.10 \\
                            & GPTQ + ReQuant   & \textbf{44.11} & \textbf{68.73} & \textbf{77.16} & 31.20          & 69.79          & \textbf{67.40} & \textbf{41.20} & 74.76          & 43.65          & \textbf{68.11} & \textbf{58.61} \\
    \cmidrule(lr){2-13}
                            & GPTAQ            & 42.92          & \textbf{69.74} & \textbf{74.46} & 32.32          & \textbf{71.26} & 66.80          & 42.40          & \textbf{75.19} & \textbf{44.42} & 67.01          & 58.65 \\
                            & GPTAQ + ReQuant  & \textbf{44.03} & 69.70          & 74.04          & \textbf{33.66} & 70.73          & \textbf{66.83} & \textbf{42.80} & 74.86          & 42.53          & \textbf{67.96} & \textbf{58.71} \\
    \cmidrule(lr){1-13}
    \multirow{4}{*}{W2A4}
                            & GPTQ             & 23.63          & 34.18          & 56.64          & \textbf{24.22} & 35.67          & 12.32          & 28.20          & 55.98          & 34.80          & 53.12          & 35.88 \\
                            & GPTQ + ReQuant   & \textbf{28.16} & \textbf{40.82} & \textbf{60.43} & 22.66          & \textbf{45.15} & \textbf{29.23} & \textbf{29.00} & \textbf{60.77} & \textbf{37.46} & \textbf{56.35} & \textbf{41.00} \\
    \cmidrule(lr){2-13}
                            & GPTAQ            & 25.94          & 39.56          & 60.95          & \textbf{25.04} & 46.17          & 27.46          & \textbf{31.00} & 61.32          & 36.49          & 55.09          & 40.90 \\
                            & GPTAQ + ReQuant  & \textbf{26.88} & \textbf{42.13} & \textbf{62.08} & 23.18          & \textbf{46.40} & \textbf{30.18} & 30.20          & \textbf{61.59} & \textbf{38.54} & \textbf{57.93} & \textbf{41.91} \\
    \bottomrule
  \end{tabular}}
\end{table}

\subsection{Effect of refinement sweeps}
\label{sec:sweep-ablation}
\begin{figure}[t]
  \centering
  \includegraphics[width=0.33\textwidth]{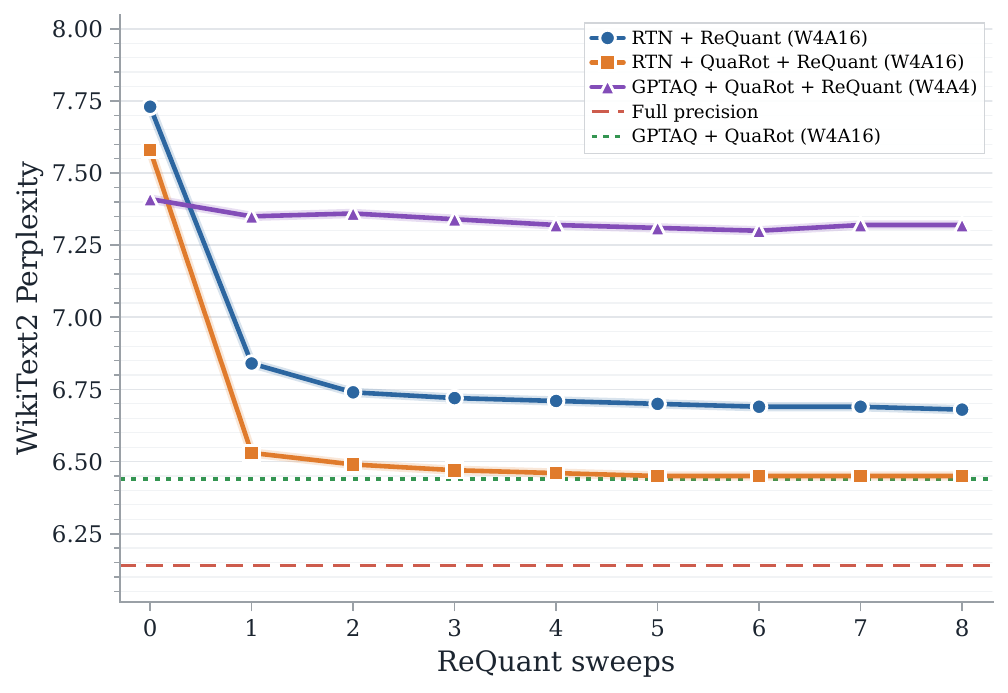}\hfill
  \includegraphics[width=0.33\textwidth]{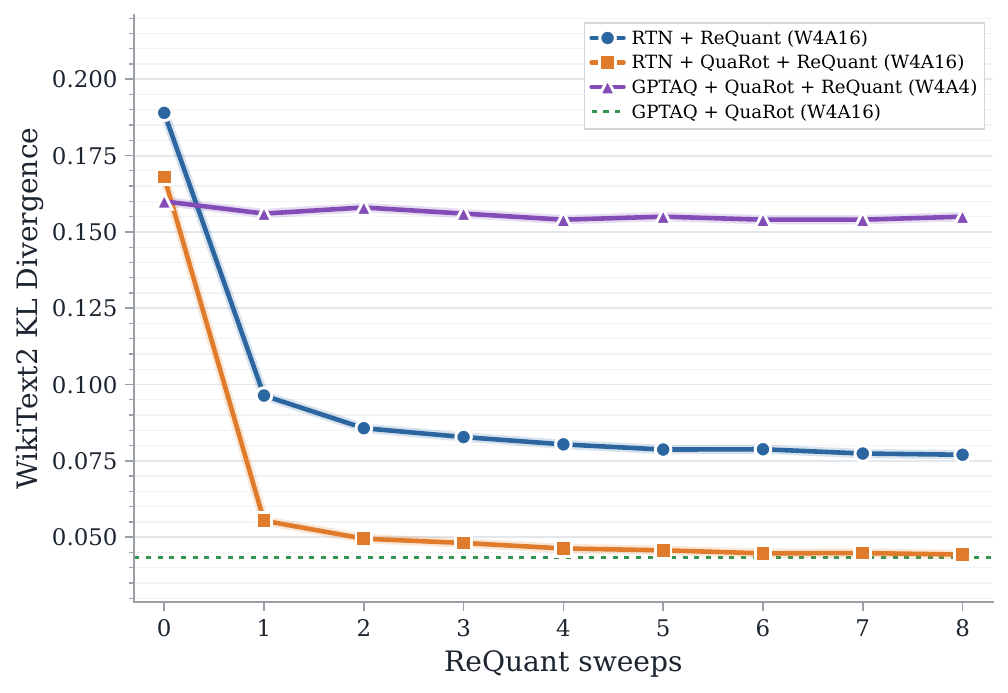}\hfill
  \includegraphics[width=0.33\textwidth]{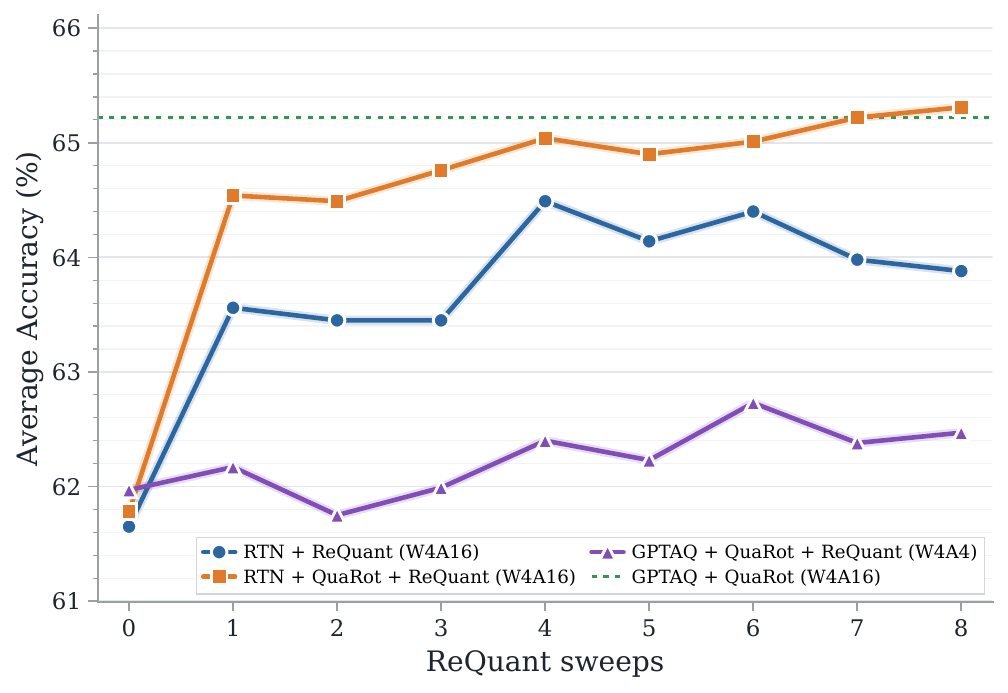}
  \caption{Effect of the number of ReQuant refinement sweeps $T$ on (left) WikiText-2 perplexity, (center) WikiText-2 KL divergence to the FP model, and (right) average zero-shot accuracy on Llama-3 8B. Three settings are compared: RTN at W4A16 (no QuaRot), RTN with QuaRot~\cite{ashkboos2024quarot} rotation at W4A16, and GPTAQ with QuaRot rotation at W4A4.}
  \label{fig:sweep-convergence}
\end{figure}
We next study how ReQuant evolves with the number of refinement sweeps $T$. Figure~\ref{fig:sweep-convergence} reports WikiText-2 perplexity, WikiText-2 KL divergence to the full-precision model, and average zero-shot accuracy on Llama-3 8B for three representative settings: RTN at W4A16 without QuaRot, RTN+QuaRot at W4A16, and GPTAQ+QuaRot at W4A4. All other settings follow Section~\ref{sec:exp-setup}. We vary $T \in \{0,1,\ldots,8\}$, where $T=0$ denotes the initial PTQ output. The PPL and KL plots also include full-precision and W4A16 GPTAQ results as horizontal references.

Three observations emerge from Figure~\ref{fig:sweep-convergence}. First, as the number of refinement sweeps $T$ increases, PPL and KL generally decrease, while average zero-shot accuracy improves, consistent with the monotone decrease of the layer-wise objective along accepted updates (Section~\ref{sec:analysis-convergence}). Second, ReQuant benefits both simple and strong initializers. For RTN at W4A16 without QuaRot, additional sweeps substantially improve average accuracy, bringing it close to the GPTAQ + QuaRot W4A16 reference value of $65.50\%$ on Llama-3 8B. For GPTAQ + QuaRot at W4A4 on the same model, refinement further raises the ten-task mean from $61.97\%$ to $62.38\%$ (Table~\ref{tab:requant-acc-main}), showing that residual fixed-grid error remains exploitable after a strong PTQ start. Third, QuaRot-based settings converge faster and achieve better final performance, suggesting that rotation makes fixed-grid optimization more favorable.

\paragraph{Cost--benefit of $T$.}
Table~\ref{tab:t-cost-benefit} reports the accuracy--runtime trade-off under W4A16 on Llama-3 8B for RTN+QuaRot and GPTQ+QuaRot with neighborhood size $K{=}2$. On RTN, $T{=}4$ uses $81.80$ minutes of end-to-end offline wall-clock; on GPTQ, $T{=}4$ uses $94.05$ minutes. Most of the held-out gain appears within the first one or two sweeps. For RTN+QuaRot, $T{=}2$ recovers a large fraction of the $T{=}8$ WikiText PPL/KL improvement at lower cost; for GPTQ+QuaRot, $T{=}2$ already reaches the best average accuracy in this sweep ($65.66\%$), and additional sweeps further reduce PPL/KL.

\begin{table}[t]
  \centering
  \caption{Cost--benefit of the number of refinement sweeps $T$ under W4A16 on Llama-3 8B ($K{=}2$). $T{=}0$ is the initializer without ReQuant. W2/UC/NM denote WikiText-2, UltraChat-2k, and NuminaMath. Time is end-to-end offline quantization wall-clock. Within each setting, the best entry in each column is bolded.}
  \label{tab:t-cost-benefit}
  \footnotesize
  \setlength{\tabcolsep}{3.5pt}
  \begin{tabular}{@{}llccccccc@{}}
    \toprule
    & & \multicolumn{2}{c}{W2} & \multicolumn{2}{c}{UC} & NM & & \\
    \cmidrule(lr){3-4}\cmidrule(lr){5-6}\cmidrule(lr){7-7}
    Setting & $T$ & KL$\downarrow$ & PPL$\downarrow$ & KL$\downarrow$ & PPL$\downarrow$ & PPL$\downarrow$ & Acc.$\uparrow$ & Time (min)$\downarrow$ \\
    \midrule
    \multirow{5}{*}{RTN+QuaRot}
      & 0 & 0.180 & 7.65 & 0.115 & 3.90 & 3.06 & 62.64 & \textbf{5.03} \\
      & 1 & 0.067 & 6.61 & 0.050 & 3.65 & 2.87 & 64.83 & 50.82 \\
      & 2 & 0.061 & 6.56 & 0.048 & 3.65 & 2.86 & 65.13 & 60.00 \\
      & 4 & 0.055 & 6.52 & 0.042 & \textbf{3.62} & \textbf{2.85} & \textbf{65.42} & 81.80 \\
      & 8 & \textbf{0.052} & \textbf{6.51} & \textbf{0.040} & \textbf{3.62} & \textbf{2.85} & 65.22 & 125.24 \\
    \cmidrule(lr){1-9}
    \multirow{5}{*}{GPTQ+QuaRot}
      & 0 & 0.054 & 6.53 & 0.038 & 3.60 & 2.85 & 65.16 & \textbf{31.62} \\
      & 1 & 0.044 & 6.44 & 0.034 & 3.59 & \textbf{2.83} & 65.41 & 60.58 \\
      & 2 & 0.043 & 6.44 & 0.033 & 3.59 & \textbf{2.83} & \textbf{65.66} & 70.86 \\
      & 4 & 0.041 & 6.43 & 0.033 & \textbf{3.58} & \textbf{2.83} & 65.33 & 94.05 \\
      & 8 & \textbf{0.040} & \textbf{6.42} & \textbf{0.032} & \textbf{3.58} & \textbf{2.83} & 65.49 & 209.57 \\
    \bottomrule
  \end{tabular}
\end{table}

\subsection{Comparison with FlexRound}
\label{sec:flexround}

To situate ReQuant relative to construction-time adaptive rounding, we compare against FlexRound~\cite{lee2023flexround} under W4A16 with the same WikiText-2 calibration budget ($512$ sequences of length $2048$) and evaluation protocol.
All methods in Table~\ref{tab:flexround} were timed on the same NVIDIA RTX PRO~6000 for a hardware-matched offline-cost comparison.
FlexRound follows its original reconstruction pipeline; ReQuant pipelines use QuaRot~\cite{ashkboos2024quarot}, which is orthogonal to ReQuant and common in low-bit Transformer PTQ.
We therefore treat Table~\ref{tab:flexround} as a \emph{pipeline-level} comparison in which metric gaps reflect the full stack (initializer, rotation, and refinement).

The paired GPTQ+QuaRot control in Appendix~\ref{app:flexround-quarot} freezes the initializer, grid, scales, and zero-points and isolates the refinement stage: PPL/KL metrics improve and average accuracy rises from $65.16\%$ to $65.33\%$.
Relative to plain FlexRound, GPTQ+QuaRot+ReQuant is stronger on both Llama-2 7B~\cite{touvron2023llama2} and Llama-3 8B in Table~\ref{tab:flexround}, with lower measured offline time.
On Llama-3 8B, RTN+QuaRot+ReQuant attains the best average accuracy among the compared pipelines ($65.42\%$), and GPTQ+QuaRot+ReQuant attains the best WikiText/UltraChat PPL--KL and NuminaMath PPL while remaining free of backpropagation and optimizer states.
These results support ReQuant as a practical backpropagation-free refinement stage within a PTQ pipeline.

\begin{table}[t]
  \centering
  \caption{Pipeline comparison with FlexRound under W4A16 (hardware-matched timing on RTX PRO~6000). ReQuant uses $T{=}8$ for Llama-2 RTN and $T{=}4$ otherwise ($K{=}2$). FlexRound+QuaRot is reported for Llama-3~8B. W2/UC/NM denote WikiText-2, UltraChat-2k, and NuminaMath. Within each model, the best entry in each column is bolded.}
  \label{tab:flexround}
  \footnotesize
  \setlength{\tabcolsep}{3.5pt}
  \begin{tabular}{@{}llccccccc@{}}
    \toprule
    & & \multicolumn{2}{c}{W2} & \multicolumn{2}{c}{UC} & NM & & \\
    \cmidrule(lr){3-4}\cmidrule(lr){5-6}\cmidrule(lr){7-7}
    Model & Method & KL$\downarrow$ & PPL$\downarrow$ & KL$\downarrow$ & PPL$\downarrow$ & PPL$\downarrow$ & Acc.$\uparrow$ & Time (min)$\downarrow$ \\
    \midrule
    \multirow{3}{*}{Llama-2 7B}
      & FlexRound & 0.034 & 5.65 & 0.029 & 3.09 & 3.26 & 60.65 & 150.67 \\
      & RTN+QuaRot+ReQuant & 0.020 & 5.59 & 0.019 & 3.06 & 3.22 & 60.69 & 95.72 \\
      & GPTQ+QuaRot+ReQuant & \textbf{0.016} & \textbf{5.57} & \textbf{0.016} & \textbf{3.05} & \textbf{3.20} & \textbf{60.96} & \textbf{70.00} \\
    \cmidrule(lr){1-9}
    \multirow{4}{*}{Llama-3 8B}
      & FlexRound & 0.075 & 6.69 & 0.067 & 3.73 & 2.97 & 65.08 & 179.50 \\
      & FlexRound+QuaRot & 0.048 & 6.47 & 0.039 & \textbf{3.58} & 2.84 & 65.24 & 188.50 \\
      & RTN+QuaRot+ReQuant & 0.055 & 6.52 & 0.042 & 3.62 & 2.85 & \textbf{65.42} & \textbf{81.80} \\
      & GPTQ+QuaRot+ReQuant & \textbf{0.041} & \textbf{6.43} & \textbf{0.033} & \textbf{3.58} & \textbf{2.83} & 65.33 & 94.05 \\
    \bottomrule
  \end{tabular}
\end{table}

\subsection{Scalability to Qwen3-235B MoE}
\label{sec:qwen235b}

Table~\ref{tab:qwen235b} evaluates W4 quantization of Qwen3-235B-A22B~\cite{yang2025qwen3} (235B total / 22B active parameters) on eight NVIDIA H200 GPUs with expert-parallel solving, QuaRot, and the same $512{\times}2048$ WikiText-2 calibration protocol used elsewhere.
Relative to GPTQ+QuaRot, four ReQuant sweeps reduce Top-20 KL on WikiText-2 / UltraChat / NuminaMath by $13.8\%$/ $13.1\%$/ $5.5\%$ and raise average accuracy from $75.00\%$ to $75.35\%$.
RTN+QuaRot+ReQuant completes in $321.27$ minutes at $T{=}4$, providing a faster offline path at this scale.
These results show that ReQuant remains effective beyond 100B parameters as a one-time offline refinement stage.

\begin{table}[t]
  \centering
  \caption{Qwen3-235B-A22B W4 scalability evaluation (8$\times$H200). W2/UC/NM denote WikiText-2, UltraChat-2k, and NuminaMath. The best entry in each column is bolded.}
  \label{tab:qwen235b}
  \footnotesize
  \setlength{\tabcolsep}{3.5pt}
  \begin{tabular}{@{}llccccccc@{}}
    \toprule
    & & \multicolumn{2}{c}{W2} & \multicolumn{2}{c}{UC} & NM & & \\
    \cmidrule(lr){3-4}\cmidrule(lr){5-6}\cmidrule(lr){7-7}
    Method & $T$ & KL$\downarrow$ & PPL$\downarrow$ & KL$\downarrow$ & PPL$\downarrow$ & PPL$\downarrow$ & Acc.$\uparrow$ & Time (min)$\downarrow$ \\
    \midrule
    GPTQ+QuaRot & 0 & 0.075 & \textbf{4.53} & 0.041 & \textbf{3.98} & \textbf{2.49} & 75.00 & 441.09 \\
    GPTQ+QuaRot+ReQuant & 4 & \textbf{0.064} & \textbf{4.53} & \textbf{0.035} & 4.12 & 2.53 & \textbf{75.35} & 695.28 \\
    RTN+QuaRot+ReQuant & 4 & 0.083 & 4.60 & 0.045 & 4.13 & 2.53 & 74.95 & \textbf{321.27} \\
    RTN+QuaRot+ReQuant & 8 & 0.079 & 4.58 & 0.044 & 4.16 & 2.52 & 74.58 & 624.92 \\
    \bottomrule
  \end{tabular}
\end{table}

\section{Conclusion}
We presented ReQuant, a fixed-grid discrete refinement stage within PTQ that treats completed PTQ outputs as feasible initializations and continues optimizing their discrete assignments. ReQuant iteratively revisits integer codes on the fixed quantization grid, accepts loss-reducing updates, and preserves the original deployment format. Across model families, bit-widths, and PTQ initializers---including a Qwen3-235B MoE setting---ReQuant improves quantized models, with especially large gains for simple initializers and lower bit-widths; the offline budget is controllable through $T$, and refinement is performed entirely offline with zero impact on serving latency.
\paragraph{Limitations.}
ReQuant performs coordinate-wise local search on a fixed quantization grid inherited from an upstream PTQ method, so its solution quality depends on the calibration set and the frozen scales/zero-points. The offline refinement cost grows with the number of sweeps $T$ and is most noticeable relative to very fast initializers such as RTN; this cost is incurred once, leaves inference unchanged, and can be reduced by choosing a smaller $T$. Like other reconstruction-based PTQ methods, ReQuant optimizes a layer-wise calibration objective, and held-out metrics are reported empirically.
\paragraph{Future Work.}
Promising directions include more efficient search strategies, extensions to joint weight-activation optimization, and stronger guarantees on approximate optimality and robustness under distribution shift.

% \section*{References}

{
\small

\bibliographystyle{unsrtnat}
\bibliography{ref}
}

%%%%%%%%%%%%%%%%%%%%%%%%%%%%%%%%%%%%%%%%%%%%%%%%%%%%%%%%%%%%

\clearpage
\appendix

\section{ReQuant derivation details}
\label{app:requant-derivations}

\subsection{ReQuant Coordinate Update Derivation}
\label{app:requant-coordinate-update}
We derive the row-wise quantities used by ReQuant and show why every candidate
move can be scored from cached statistics. For one output row, let
\(\mathbf w\) be the full-precision row, \(\mathbf q\) the current quantized
row, and \(\mathbf e:=\mathbf w-\mathbf q\) the quantization error. With
\(\Delta\mathbf X:=\widetilde{\mathbf X}-\mathbf X\), the activation-aware
row loss in Eq.~\ref{eq:row_objective} can be rewritten as
\[
  \begin{aligned}
  L(\mathbf e)
  &=
  \|\mathbf w\mathbf X-\mathbf q\widetilde{\mathbf X}\|_2^2  \\
  &=
  \|\mathbf w(\mathbf X-\widetilde{\mathbf X})
    +(\mathbf w-\mathbf q)\widetilde{\mathbf X}\|_2^2 \\
  &=
  \|\mathbf e\widetilde{\mathbf X}-\mathbf w\Delta\mathbf X\|_2^2 .
  \end{aligned}
\]
Thus \(L\) is a quadratic function of \(\mathbf e\). Defining
\[
  \widetilde{\mathbf H}:=\widetilde{\mathbf X}\widetilde{\mathbf X}^{\top},
  \qquad
  \mathbf B:=\Delta\mathbf X\widetilde{\mathbf X}^{\top},
\]
and expanding the squared norm gives
\[
  L(\mathbf e)
  =
  \mathbf e\widetilde{\mathbf H}\mathbf e^\top
  -2\mathbf e\mathbf B^\top\mathbf w^\top
  +\mathbf w\Delta\mathbf X\Delta\mathbf X^\top\mathbf w^\top .
\]
The last term is constant with respect to \(\mathbf e\), so the gradient vector
used in the main text is
\[
  \mathbf g
  :=
  \nabla_{\mathbf e}L(\mathbf e)
  =
  2\bigl(\mathbf e\widetilde{\mathbf H}-\mathbf w\mathbf B\bigr).
\]

Now consider changing coordinate \(j\) of the quantized row by an on-grid step
\(\Delta q_j\). Since \(\mathbf e=\mathbf w-\mathbf q\), increasing \(q_j\) by
\(\Delta q_j\) changes the error to
\[
  \mathbf e'=\mathbf e-\Delta q_j\mathbf u_j ,
\]
where \(\mathbf u_j\) is the \(j\)-th unit vector. The exact loss change is
\[
  \begin{aligned}
  \Delta L(\Delta q_j)
  &:=L(\mathbf e')-L(\mathbf e) \\
  &=
  \|(\mathbf e-\Delta q_j\mathbf u_j)\widetilde{\mathbf X}
    -\mathbf w\Delta\mathbf X\|_2^2
  -\|\mathbf e\widetilde{\mathbf X}-\mathbf w\Delta\mathbf X\|_2^2 \\
  &=
  -2\Delta q_j\,
  \mathbf u_j\widetilde{\mathbf X}
  (\mathbf e\widetilde{\mathbf X}-\mathbf w\Delta\mathbf X)^\top
  +(\Delta q_j)^2
  \|\mathbf u_j\widetilde{\mathbf X}\|_2^2 \\
  &=
  -2\Delta q_j\,(\mathbf e\widetilde{\mathbf H}-\mathbf w\mathbf B)_j
  +(\Delta q_j)^2\widetilde H_{jj} \\
  &=
  -\Delta q_j\,g_j
  +(\Delta q_j)^2\widetilde H_{jj}.
  \end{aligned}
\]
This proves Eq.~\ref{eq:coordinate_delta}. The key point is that evaluating a
candidate move requires only \(g_j\) and \(\widetilde H_{jj}\), not a
recomputation of the row residual.

The feasible candidates are determined by the integer quantization code. If
\(q_j=(z_j-o_j)s_j\), any valid move has the form
\(\Delta q_j=k s_j\), where \(k\in\mathbb Z\setminus\{0\}\) and
\(z_{\min}\le z_j+k\le z_{\max}\). Therefore the best coordinate move is the
finite enumeration in Eq.~\ref{eq:best_move}. After an accepted move, the state
maintained by ReQuant is updated exactly. The error update is immediate from
\(\mathbf e'=\mathbf e-\Delta q_j^\star\mathbf u_j\), and the gradient vector
satisfies
\[
  \begin{aligned}
  \mathbf g'
  &=
  2\bigl(\mathbf e'\widetilde{\mathbf H}-\mathbf w\mathbf B\bigr) \\
  &=
  2\bigl((\mathbf e-\Delta q_j^\star\mathbf u_j)\widetilde{\mathbf H}
  -\mathbf w\mathbf B\bigr) \\
  &=
  \mathbf g-2\Delta q_j^\star\widetilde{\mathbf H}_{j,:}.
  \end{aligned}
\]
Finally, because Eq.~\ref{eq:coordinate_delta} gives the exact loss difference,
the maintained scalar loss is updated by
\(L\leftarrow L+\Delta L(\Delta q_j^\star)\). Together these identities justify
the incremental updates used in Algorithm~\ref{alg:refine}.

\subsection{Finite Termination of ReQuant}
\label{app:requant-termination}

We prove finite termination for one row under fixed calibration statistics and a
fixed quantization grid; the layer-wise result follows from the row
decomposition in Eq.~\ref{eq:row-decomposition}. Let the feasible values of
coordinate \(j\) be the finite set \(\mathcal Q_j\). Under \(b\)-bit
quantization, \(|\mathcal Q_j|\le 2^b\). The feasible set for one quantized row
is therefore the Cartesian product
\[
  \mathcal Q
  =
  \mathcal Q_1\times \mathcal Q_2\times\cdots\times
  \mathcal Q_{d_{\mathrm{col}}},
\]
whose cardinality satisfies
\[
  |\mathcal Q|
  =
  \prod_{j=1}^{d_{\mathrm{col}}}|\mathcal Q_j|
  \le
  (2^b)^{d_{\mathrm{col}}}
  =
  2^{b d_{\mathrm{col}}}.
\]
Hence \(\mathcal Q\) is finite.

Let \(\mathbf q^{(r)}\in\mathcal Q\) denote the quantized row after the
\(r\)-th accepted update, and let
\[
  \mathbf e^{(r)}=\mathbf w-\mathbf q^{(r)}
\]
be the corresponding quantization error. ReQuant accepts an update only if
\[
  \Delta L^{(r)}
  :=
  L(\mathbf e^{(r+1)})-L(\mathbf e^{(r)})
  <
  0 .
\]
Therefore every accepted update strictly decreases the row objective:
\[
  L(\mathbf e^{(0)})
  >
  L(\mathbf e^{(1)})
  >
  L(\mathbf e^{(2)})
  >
  \cdots .
\]

We now prove by contradiction that infinitely many accepted updates are
impossible. Suppose ReQuant accepted infinitely many updates. Then it would
generate an infinite sequence
\[
  \mathbf q^{(0)},\mathbf q^{(1)},\mathbf q^{(2)},\ldots
\]
with every \(\mathbf q^{(r)}\in\mathcal Q\). Since \(\mathcal Q\) is finite, any
infinite sequence in \(\mathcal Q\) must repeat an element. Hence there exist two
indices \(a<b\) such that
\[
  \mathbf q^{(a)}=\mathbf q^{(b)} .
\]
Because \(\mathbf e=\mathbf w-\mathbf q\), this equality also implies
\[
  \mathbf e^{(a)}
  =
  \mathbf w-\mathbf q^{(a)}
  =
  \mathbf w-\mathbf q^{(b)}
  =
  \mathbf e^{(b)} ,
\]
and thus
\[
  L(\mathbf e^{(a)})=L(\mathbf e^{(b)}) .
\]
However, the sequence contains at least one accepted update between \(a\) and
\(b\), and every accepted update strictly decreases \(L\). Hence
\[
  L(\mathbf e^{(b)})<L(\mathbf e^{(a)}),
\]
which contradicts \(L(\mathbf e^{(a)})=L(\mathbf e^{(b)})\).

Thus ReQuant cannot accept infinitely many updates. Since each accepted update
visits a new quantized row, the number of accepted updates is at most
\[
  |\mathcal Q|-1
  \le
  2^{b d_{\mathrm{col}}}-1 .
\]
This bound is extremely loose, but it is sufficient to establish finite
termination under fixed calibration statistics.

\subsection{Complexity Analysis}
\label{app:requant-complexity}

We analyze the cost of ReQuant for one linear layer with
\(\mathbf W\in\mathbb R^{d_{\mathrm{row}}\times d_{\mathrm{col}}}\) and
\(\mathbf X,\widetilde{\mathbf X}\in\mathbb R^{d_{\mathrm{col}}\times m}\),
where \(m\) is the number of calibration tokens. Let \(K\) denote the number of
candidate grid moves considered per coordinate. If all feasible nonzero moves
are enumerated for \(b\)-bit quantization, then \(K\le 2^b-1\); if only nearby
grid points are considered, \(K\) is a small constant.

\paragraph{Naive candidate evaluation.}
The most direct way to test whether changing one entry \(W_{q,i,j}\) improves
the objective is to recompute the full layer reconstruction loss
\(\|\mathbf W\mathbf X-\mathbf W_q\widetilde{\mathbf X}\|_F^2\). This requires
a matrix multiplication and costs \(O(m d_{\mathrm{row}}d_{\mathrm{col}})\) per
candidate. Since a layer contains \(d_{\mathrm{row}}d_{\mathrm{col}}\)
coordinates and each coordinate has \(K\) candidates, naive exhaustive
evaluation would cost
\[
  O(K\,m\,d_{\mathrm{row}}^2d_{\mathrm{col}}^2),
\]
which is prohibitive for LLM layers.

\paragraph{Effect of row-wise decomposition.}
By Eq.~\ref{eq:row-decomposition}, changing \(W_{q,i,j}\) affects only the
\(i\)-th row loss
\[
  L_i
  =
  \|\mathbf W_{i,:}\mathbf X-\mathbf W_{q,i,:}\widetilde{\mathbf X}\|_2^2 .
\]
If this row loss were recomputed from scratch for each candidate, the cost would
be \(O(m d_{\mathrm{col}})\) per candidate, reducing the full-layer factor
\(d_{\mathrm{row}}\) and enabling all rows to be processed independently. This
is still expensive because it repeatedly accesses the calibration activations.

\paragraph{Closed-form candidate scoring.}
ReQuant instead uses the quadratic form in Eq.~\ref{eq:row_objective}. Once
\(\mathbf g=2(\mathbf e\widetilde{\mathbf H}-\mathbf w\mathbf B)\) is available,
Eq.~\ref{eq:coordinate_delta} scores one candidate using only
\(\Delta q_j\), \(g_j\), and \(\widetilde H_{jj}\), so each candidate costs
\(O(1)\). Scanning all \(K\) candidates at one coordinate costs \(O(K)\); scanning
all coordinates of all rows in one sweep costs
\[
  O(d_{\mathrm{row}}d_{\mathrm{col}}K).
\]

\paragraph{Accepted-update maintenance.}
When a move is accepted, ReQuant refreshes the gradient vector by
\(\mathbf g\leftarrow \mathbf g-2\Delta q_j^\star\widetilde{\mathbf H}_{j,:}\).
This costs \(O(d_{\mathrm{col}})\), since \(\widetilde{\mathbf H}_{j,:}\) has
length \(d_{\mathrm{col}}\). Recomputing
\(\mathbf g=2(\mathbf e\widetilde{\mathbf H}-\mathbf w\mathbf B)\) from scratch for
one row would cost \(O(d_{\mathrm{col}}^2)\). If \(A\) moves are accepted in one
sweep, the state-maintenance cost is therefore \(O(A d_{\mathrm{col}})\). In the
worst case, \(A\le d_{\mathrm{row}}d_{\mathrm{col}}\), giving
\(O(d_{\mathrm{row}}d_{\mathrm{col}}^2)\) per sweep for accepted-update
maintenance.

\paragraph{Precomputation and initialization.}
Before refinement, ReQuant computes
\[
  \widetilde{\mathbf H}
  =
  \widetilde{\mathbf X}\widetilde{\mathbf X}^{\top},
  \qquad
  \mathbf B
  =
  \Delta\mathbf X\widetilde{\mathbf X}^{\top},
\]
which costs \(O(m d_{\mathrm{col}}^2)\) and stores
\(O(d_{\mathrm{col}}^2)\) statistics shared by all output rows. Initializing the
gradient matrix for all rows,
\[
  \mathbf G
  =
  2(\mathbf E\widetilde{\mathbf H}-\mathbf W\mathbf B),
  \qquad
  \mathbf E=\mathbf W-\mathbf Q,
\]
costs \(O(d_{\mathrm{row}}d_{\mathrm{col}}^2)\). Initializing all row losses
from residuals costs \(O(m d_{\mathrm{row}}d_{\mathrm{col}})\), which is usually
dominated by the matrix-statistic and gradient-initialization terms.

\paragraph{Total cost.}
For \(T\) refinement sweeps and average accepted updates \(A_{\mathrm{avg}}\) per
sweep, the layer-wise ReQuant complexity is
\[
  O(m d_{\mathrm{col}}^2)
  +
  O(d_{\mathrm{row}}d_{\mathrm{col}}^2)
  +
  O(T d_{\mathrm{row}}d_{\mathrm{col}}K)
  +
  O(T A_{\mathrm{avg}}d_{\mathrm{col}}).
\]
Using \(A_{\mathrm{avg}}\le d_{\mathrm{row}}d_{\mathrm{col}}\), the worst-case
bound becomes
\[
  O(m d_{\mathrm{col}}^2)
  +
  O((T+1)d_{\mathrm{row}}d_{\mathrm{col}}^2)
  +
  O(Td_{\mathrm{row}}d_{\mathrm{col}}K).
\]
For low-bit quantization \(K\le 2^b-1\) and typically
\(K\ll d_{\mathrm{col}}\), so the dominant terms are
\[
  O(m d_{\mathrm{col}}^2 + (T+1)d_{\mathrm{row}}d_{\mathrm{col}}^2).
\]

%%%%%%%%%%%%%%%%%%%%%%%%%%%%%%%%%%%%%%%%%%%%%%%%%%%%%%%%%%%%

\clearpage
\section{Additional experimental results}
\label{app:additional-experiments}

This appendix collects supplementary experimental results that did not fit in the main text. All \emph{KL/PPL} tables below are evaluated on WikiText-2, UltraChat-2k, and NuminaMath; we abbreviate the three column groups as \textbf{W2}, \textbf{UC}, and \textbf{NM}, matching the headers. \textbf{KL} is the (unscaled) KL divergence to the full-precision model on each evaluation text. \textbf{PPL} is standard \emph{token-level} perplexity on the same text. Unless noted otherwise, we report PPL to two decimal places and KL to three. Lower is better for both. Unless noted otherwise, \textbf{within each baseline pair} (each PTQ initializer vs.\ the same initializer after ReQuant), \textbf{the better entry is bolded.} Section~\ref{app:kl-ppl-details} reports these metrics for Llama-3 8B and Qwen3-14B at W4A16. Section~\ref{app:kl-ppl-w4a4-8b-14b} reports them at W4A4 for the same models. Section~\ref{app:kl-ppl-70b} reports them on Llama-3 70B at W4A4. Section~\ref{app:w4a16-no-rotation} ablates QuaRot by re-running the W4A16 experiments without any orthogonal preprocessing. Section~\ref{app:runtime-qwen} reports the runtime overhead on Qwen3-14B. Section~\ref{app:low-bit-kl-ppl} reports KL/PPL for the W3A4/W2A4 setting in Section~\ref{sec:low-bit}. Section~\ref{app:gptaq-calibration-samples} plots WikiText-2 PPL/KL and average downstream accuracy as a function of calibration sample size for GPTAQ with QuaRot and ReQuant at W4A4 on Llama-3 8B. Sections~\ref{app:flexround-quarot}--\ref{app:search-ablations} add FlexRound/QuaRot controls and search-procedure ablations ($K$, coordinate order, seed). The Qwen3-235B MoE scalability study is reported in Section~\ref{sec:qwen235b}.

\subsection{W4A16 results on Llama-3 8B and Qwen3-14B}
\label{app:kl-ppl-details}

At W4A16 under QuaRot and per-weight--activation quantization, Table~\ref{tab:requant-w4a16-kl-ppl} shows that ReQuant reduces both KL divergence and perplexity versus each initializer-alone baseline across Llama-3 8B and Qwen3-14B and across W2/UC/NM.
The KL/PPL deltas are especially large for simple initial grids (especially RTN, then AWQ), and GPTAQ still exhibits consistent KL/PPL improvements in most bins.

\begin{table}[ht]
  \centering
  \caption{KL and PPL under W4A16 quantization on Llama-3 8B and Qwen3-14B with QuaRot~\cite{ashkboos2024quarot} rotation.}
  \label{tab:requant-w4a16-kl-ppl}
  \scriptsize
  \setlength{\tabcolsep}{1.6pt}
  \resizebox{0.92\textwidth}{!}{%
  \begin{tabular}{@{}l|*{6}{c}|*{6}{c}@{}}
    \toprule
    \multicolumn{1}{@{}l|}{} & \multicolumn{6}{c|}{Llama-3 8B} & \multicolumn{6}{c@{}}{Qwen3-14B} \\
    \cmidrule(lr){2-7}\cmidrule(lr){8-13}
    \multicolumn{1}{@{}l|}{}
    & \multicolumn{2}{c}{W2} & \multicolumn{2}{c}{UC} & \multicolumn{2}{c|}{NM}
    & \multicolumn{2}{c}{W2} & \multicolumn{2}{c}{UC} & \multicolumn{2}{c@{}}{NM} \\
    \cmidrule(lr){2-3}\cmidrule(lr){4-5}\cmidrule(lr){6-7}
    \cmidrule(lr){8-9}\cmidrule(lr){10-11}\cmidrule(lr){12-13}
    Method
    & KL$\downarrow$ & PPL$\downarrow$ & KL$\downarrow$ & PPL$\downarrow$ & KL$\downarrow$ & PPL$\downarrow$
    & KL$\downarrow$ & PPL$\downarrow$ & KL$\downarrow$ & PPL$\downarrow$ & KL$\downarrow$ & PPL$\downarrow$ \\
    \midrule
    FP                & --             & 6.14           & --             & 3.47           & --             & 2.71           & --             & 8.65           & --             & 4.98           & --             & 3.36 \\
    \midrule
    RTN               & 0.174          & 7.63           & 0.116          & 3.94           & 0.100          & 3.05           & 0.176          & 10.22          & 0.145          & 5.20           & 0.102          & \textbf{3.27} \\
    RTN + ReQuant     & \textbf{0.079} & \textbf{6.71}  & \textbf{0.063} & \textbf{3.70}  & \textbf{0.068} & \textbf{2.91}  & \textbf{0.031} & \textbf{8.85}  & \textbf{0.036} & \textbf{4.98}  & \textbf{0.029} & 3.33 \\
    \cmidrule(lr){1-13}
    AWQ               & 0.091          & 6.82           & 0.064          & 3.69           & 0.054          & \textbf{2.87}  & 0.140          & 9.68           & 0.117          & 5.16           & 0.078          & 3.43 \\
    AWQ + ReQuant     & \textbf{0.069} & \textbf{6.65}  & \textbf{0.051} & \textbf{3.65}  & \textbf{0.053} & \textbf{2.87}  & \textbf{0.052} & \textbf{8.96}  & \textbf{0.054} & \textbf{5.04}  & \textbf{0.036} & \textbf{3.39} \\
    \cmidrule(lr){1-13}
    GPTQ              & 0.054          & 6.53           & 0.038          & \textbf{3.59}  & \textbf{0.041} & \textbf{2.83}  & 0.027          & \textbf{8.78}  & \textbf{0.031} & 4.97           & \textbf{0.024} & \textbf{3.34} \\
    GPTQ + ReQuant    & \textbf{0.049} & \textbf{6.49}  & \textbf{0.036} & \textbf{3.59}  & \textbf{0.041} & 2.84           & \textbf{0.025} & 8.81           & \textbf{0.031} & \textbf{4.94}  & 0.025          & 3.36 \\
    \cmidrule(lr){1-13}
    GPTAQ             & 0.043          & 6.44           & 0.032          & 3.59           & 0.040          & 2.84           & 0.027          & 8.83           & 0.033          & 4.96           & \textbf{0.026} & 3.36 \\
    GPTAQ + ReQuant   & \textbf{0.040} & \textbf{6.41}  & \textbf{0.031} & \textbf{3.58}  & \textbf{0.028} & \textbf{2.83}  & \textbf{0.026} & \textbf{8.80}  & \textbf{0.033} & \textbf{4.91}  & \textbf{0.026} & \textbf{3.34} \\
    \bottomrule
  \end{tabular}}
\end{table}

\subsection{W4A4 results on Llama-3 8B and Qwen3-14B}
\label{app:kl-ppl-w4a4-8b-14b}

Under W4A4, Table~\ref{tab:requant-kl-ppl-main} records much larger raw KL and PPL than at W4A16 for the same models, reflecting the added difficulty of quantizing activations.
ReQuant still improves almost every baseline pair on every corpus, with the same qualitative pattern as in the weight-only regime: naive initializers incur the largest distortions relative to FP, leaving the biggest room for refinement, whereas GPTAQ starts from comparatively low divergence and perplexity.

\begin{table}[ht]
  \centering
  \caption{KL and PPL under W4A4 quantization on Llama-3 8B and Qwen3-14B with QuaRot~\cite{ashkboos2024quarot} rotation.}
  \label{tab:requant-kl-ppl-main}
  \scriptsize
  \setlength{\tabcolsep}{1.6pt}
  \resizebox{0.92\textwidth}{!}{%
  \begin{tabular}{@{}l|*{6}{c}|*{6}{c}@{}}
    \toprule
    \multicolumn{1}{@{}l|}{} & \multicolumn{6}{c|}{Llama-3 8B} & \multicolumn{6}{c@{}}{Qwen3-14B} \\
    \cmidrule(lr){2-7}\cmidrule(lr){8-13}
    \multicolumn{1}{@{}l|}{}
    & \multicolumn{2}{c}{W2} & \multicolumn{2}{c}{UC} & \multicolumn{2}{c|}{NM}
    & \multicolumn{2}{c}{W2} & \multicolumn{2}{c}{UC} & \multicolumn{2}{c@{}}{NM} \\
    \cmidrule(lr){2-3}\cmidrule(lr){4-5}\cmidrule(lr){6-7}
    \cmidrule(lr){8-9}\cmidrule(lr){10-11}\cmidrule(lr){12-13}
    Method
    & KL$\downarrow$ & PPL$\downarrow$ & KL$\downarrow$ & PPL$\downarrow$ & KL$\downarrow$ & PPL$\downarrow$
    & KL$\downarrow$ & PPL$\downarrow$ & KL$\downarrow$ & PPL$\downarrow$ & KL$\downarrow$ & PPL$\downarrow$ \\
    \midrule
    FP                & -- & 6.14  & -- & 3.47 & -- & 2.71 & -- & 8.65  & -- & 4.98 & -- & 3.36 \\
    RTN               & 0.538 & 13.21 & 0.401 & 5.71 & 0.304 & 3.89 & 0.322 & 11.44 & 0.255 & 5.37 & 0.186 & \textbf{3.41} \\
    RTN + ReQuant     & \textbf{0.165} & \textbf{7.38} & \textbf{0.130} & \textbf{3.97} & \textbf{0.122} & \textbf{3.10} & \textbf{0.149} & \textbf{9.69} & \textbf{0.138} & \textbf{5.10} & \textbf{0.099} & 3.46 \\
    \cmidrule(lr){1-13}
    AWQ               & 0.408 & 10.99 & 0.312 & 5.08 & 0.220 & 2.96 & 0.271 & 10.43 & 0.224 & 5.24 & 0.153 & \textbf{3.42} \\
    AWQ + ReQuant     & \textbf{0.221} & \textbf{7.94} & \textbf{0.166} & \textbf{4.14} & \textbf{0.115} & \textbf{2.56} & \textbf{0.165} & \textbf{9.80} & \textbf{0.150} & \textbf{5.12} & \textbf{0.101} & 3.48 \\
    \cmidrule(lr){1-13}
    GPTQ              & 0.190 & 7.86  & 0.143 & 4.07 & 0.121 & 3.12 & 0.155 & 9.71  & 0.131 & \textbf{5.03} & \textbf{0.088} & \textbf{3.37} \\
    GPTQ + ReQuant    & \textbf{0.167} & \textbf{7.46} & \textbf{0.122} & \textbf{3.95} & \textbf{0.113} & \textbf{3.08} & \textbf{0.143} & \textbf{9.61} & \textbf{0.128} & 5.06 & 0.089 & 3.43 \\
    \cmidrule(lr){1-13}
    GPTAQ             & 0.162 & 7.40  & 0.115 & 3.93 & \textbf{0.110} & \textbf{3.06} & 0.158 & 9.73  & 0.149 & \textbf{5.03} & 0.106 & 3.47 \\
    GPTAQ + ReQuant   & \textbf{0.158} & \textbf{7.36} & \textbf{0.113} & \textbf{3.91} & 0.112 & 3.07 & \textbf{0.152} & \textbf{9.67} & \textbf{0.145} & 5.06 & \textbf{0.101} & \textbf{3.42} \\
    \bottomrule
  \end{tabular}}
\end{table}

\subsection{KL/PPL under W3A4 and W2A4 on Llama-3 8B}
\label{app:low-bit-kl-ppl}

Table~\ref{tab:requant-low-bit-kl-ppl} reports WikiText-2, UltraChat-2k, and NuminaMath KL divergence and perplexity for the low-bit setting in Section~\ref{sec:low-bit}. The corresponding zero-shot accuracies appear in Table~\ref{tab:requant-low-bit}.

\begin{table}[ht]
  \centering
  \caption{KL divergence and perplexity (PPL) on Llama-3 8B at W3A4 and W2A4 with QuaRot~\cite{ashkboos2024quarot} rotation, evaluated on WikiText-2 (W2), UltraChat-2k (UC), and NuminaMath (NM).}
  \label{tab:requant-low-bit-kl-ppl}
  \scriptsize
  \setlength{\tabcolsep}{2.5pt}
  \begin{tabular}{@{}c|l|cc|cc|cc@{}}
    \toprule
    \multicolumn{2}{@{}l|}{}
    & \multicolumn{2}{c|}{W2} & \multicolumn{2}{c|}{UC} & \multicolumn{2}{c@{}}{NM} \\
    \cmidrule(lr){3-4}\cmidrule(lr){5-6}\cmidrule(lr){7-8}
    Precision & Method
    & KL$\downarrow$ & PPL$\downarrow$ & KL$\downarrow$ & PPL$\downarrow$ & KL$\downarrow$ & PPL$\downarrow$ \\
    \midrule
    \multirow{4}{*}{W3A4}
                          & GPTQ             & 0.305          & 9.21           & 0.269          & 4.74           & 0.245          & 3.60  \\
                          & GPTQ + ReQuant   & \textbf{0.251} & \textbf{8.19}  & \textbf{0.220} & \textbf{4.38}  & \textbf{0.216} & \textbf{3.47} \\
    \cmidrule(lr){2-8}
                          & GPTAQ            & 0.253          & 8.27           & \textbf{0.212} & 4.35           & 0.211          & 3.42  \\
                          & GPTAQ + ReQuant  & \textbf{0.242} & \textbf{8.13}  & 0.213          & 4.35           & \textbf{0.207} & \textbf{3.41} \\
    \cmidrule(lr){1-8}
    \multirow{4}{*}{W2A4}
                          & GPTQ             & 1.090          & 31.98          & 1.180          & 21.57          & 1.100          & 12.82 \\
                          & GPTQ + ReQuant   & \textbf{0.735} & \textbf{15.06} & \textbf{0.838} & \textbf{9.71}  & \textbf{0.995} & \textbf{9.10} \\
    \cmidrule(lr){2-8}
                          & GPTAQ            & 0.743          & 15.49          & 0.811          & 9.57           & 1.050          & 10.18 \\
                          & GPTAQ + ReQuant  & \textbf{0.701} & \textbf{14.38} & \textbf{0.789} & \textbf{9.15}  & \textbf{1.040} & \textbf{9.81} \\
    \bottomrule
  \end{tabular}
\end{table}

%%%%%%%%%%%%%%%%%%%%%%%%%%%%%%%%%%%%%%%%%%%%%%%%%%%%%%%%%%%%

\subsection{Runtime results on Qwen3-14B}
\label{app:runtime-qwen}

Table~\ref{tab:runtime-qwen} reports the end-to-end quantization time on Qwen3-14B for different PTQ initializers with QuaRot, both with and without ReQuant. For example, RTN+QuaRot uses $15.6$ minutes and RTN+QuaRot+ReQuant uses $117.1$ minutes; GPTAQ+QuaRot uses $35.3$ minutes and GPTAQ+QuaRot+ReQuant uses $128.6$ minutes. The offline budget is controllable through $T$ (Section~\ref{sec:sweep-ablation}).

This cost is incurred once during offline model preparation and leaves the deployed representation and inference latency unchanged. In our implementation the overhead is mainly computational, because ReQuant stores neither gradients nor optimizer states for learnable quantizer parameters.

\begin{table}[ht]
  \centering
  \caption{Runtime on Qwen3-14B.}
  \label{tab:runtime-qwen}
  \footnotesize
  \setlength{\tabcolsep}{8pt}
  \begin{tabular}{@{}lc@{}}
    \toprule
    Method & Runtime \\
    \midrule
    RTN + QuaRot & 15.6 min \\
    RTN + QuaRot + ReQuant & 117.1 min \\
    \cmidrule(lr){1-2}
    AWQ + QuaRot & 23.9 min \\
    AWQ + QuaRot + ReQuant & 109.8 min \\
    \cmidrule(lr){1-2}
    GPTQ + QuaRot & 31.4 min \\
    GPTQ + QuaRot + ReQuant & 128.6 min \\
    \cmidrule(lr){1-2}
    GPTAQ + QuaRot & 35.3 min \\
    GPTAQ + QuaRot + ReQuant & 128.6 min \\
    \bottomrule
  \end{tabular}
\end{table}

%%%%%%%%%%%%%%%%%%%%%%%%%%%%%%%%%%%%%%%%%%%%%%%%%%%%%%%%%%%%

\subsection{KL divergence and perplexity results on Llama-3 70B}
\label{app:kl-ppl-70b}

Table~\ref{tab:requant-70b-kl-ppl} focuses on Llama-3 70B at W4A4 under QuaRot~\cite{ashkboos2024quarot}.
RTN alone exhibits extreme KL/PPL on W2 relative to FP, and ReQuant closes most of that gap while also improving AWQ.
For GPTQ and GPTAQ the initial mismatch is already smaller; ReQuant still tends to help on perplexity and many KL cells, though a few entries favor the unrefined baseline, highlighting that fixed-grid moves can redistribute error across corpora at large scale.

\begin{table}[ht]
  \centering
  \caption{KL and PPL under W4A4 quantization on Llama-3 70B with QuaRot~\cite{ashkboos2024quarot} rotation.}
  \label{tab:requant-70b-kl-ppl}
  \footnotesize
  \setlength{\tabcolsep}{6pt}
  \begin{tabular}{@{}l|cc|cc|cc@{}}
    \toprule
    \multicolumn{1}{@{}l|}{}
    & \multicolumn{2}{c|}{W2} & \multicolumn{2}{c|}{UC} & \multicolumn{2}{c@{}}{NM} \\
    \cmidrule(lr){2-3}\cmidrule(lr){4-5}\cmidrule(lr){6-7}
    Method
    & KL$\downarrow$ & PPL$\downarrow$ & KL$\downarrow$ & PPL$\downarrow$ & KL$\downarrow$ & PPL$\downarrow$ \\
    \midrule
    FP               & --             & 2.86           & --             & 3.07           & --             & 2.40 \\
    \midrule
    RTN              & 1.953          & 60.14          & 1.405          & 27.40          & 1.241          & 13.34 \\
    RTN + ReQuant    & \textbf{0.678} & \textbf{6.00}  & \textbf{0.126} & \textbf{3.49}  & \textbf{0.143} & \textbf{2.76} \\
    \cmidrule(lr){1-7}
    AWQ              & 1.158          & 14.26          & 0.570          & 6.33           & 0.550          & 4.58 \\
    AWQ + ReQuant    & \textbf{0.683} & \textbf{6.07}  & \textbf{0.145} & \textbf{3.56}  & \textbf{0.160} & \textbf{2.80} \\
    \cmidrule(lr){1-7}
    GPTQ             & 0.753          & 7.63           & 0.307          & 4.41           & 0.269          & 3.21 \\
    GPTQ + ReQuant   & \textbf{0.743} & \textbf{6.59}  & \textbf{0.168} & \textbf{3.65}  & \textbf{0.182} & \textbf{2.87} \\
    \cmidrule(lr){1-7}
    GPTAQ            & \textbf{0.693} & 6.33           & 0.130          & 3.51           & 0.154          & 2.80 \\
    GPTAQ + ReQuant  & 0.707          & \textbf{6.27}  & \textbf{0.127} & \textbf{3.50}  & \textbf{0.148} & \textbf{2.78} \\
    \bottomrule
  \end{tabular}
\end{table}

\subsection{W4A16 results without QuaRot rotation}
\label{app:w4a16-no-rotation}

Tables~\ref{tab:requant-w4a16-kl-ppl-norot} and~\ref{tab:requant-w4a16-acc-norot} repeat the Llama-3 8B W4A16 study \emph{without} QuaRot orthogonal preprocessing.
On KL/PPL, ReQuant tends to lower both metrics relative to each initializer alone across W2/UC/NM, matching the qualitative pattern under rotation but with a few small reversals on individual cells.
On downstream accuracy (mean over ten tasks), ReQuant improves RTN ($61.65 \rightarrow 63.20$, $+1.55$), AWQ ($63.50 \rightarrow 64.06$), and GPTAQ ($64.28 \rightarrow 64.68$), while for GPTQ the average drops slightly ($64.63 \rightarrow 64.41$) even though its WikiText-2 KL/PPL still improve---illustrating that fixed-grid refinement can shift the accuracy--perplexity trade-off when rotation is absent.
\begin{table}[ht]
  \centering
  \caption{KL and PPL under W4A16 on Llama-3 8B \emph{without} QuaRot rotation.}
  \label{tab:requant-w4a16-kl-ppl-norot}
  \footnotesize
  \setlength{\tabcolsep}{6pt}
  \begin{tabular}{@{}l|cc|cc|cc@{}}
    \toprule
    \multicolumn{1}{@{}l|}{}
    & \multicolumn{2}{c|}{W2} & \multicolumn{2}{c|}{UC} & \multicolumn{2}{c@{}}{NM} \\
    \cmidrule(lr){2-3}\cmidrule(lr){4-5}\cmidrule(lr){6-7}
    Method
    & KL$\downarrow$ & PPL$\downarrow$ & KL$\downarrow$ & PPL$\downarrow$ & KL$\downarrow$ & PPL$\downarrow$ \\
    \midrule
    FP                & --             & 6.14           & --             & 3.47           & --             & 2.71 \\
    \midrule
    RTN               & 0.255          & 8.53           & 0.218          & 4.42           & 0.190          & 3.37 \\
    RTN + ReQuant     & \textbf{0.103} & \textbf{6.91}  & \textbf{0.091} & \textbf{3.81}  & \textbf{0.131} & \textbf{3.15} \\
    \cmidrule(lr){1-7}
    AWQ               & 0.115          & 7.06           & 0.080          & 3.77           & \textbf{0.079} & \textbf{2.96} \\
    AWQ + ReQuant     & \textbf{0.082} & \textbf{6.75}  & \textbf{0.070} & \textbf{3.72}  & 0.083          & 2.97 \\
    \cmidrule(lr){1-7}
    GPTQ              & 0.090          & 6.81           & \textbf{0.092} & \textbf{3.82}  & \textbf{0.100} & \textbf{3.03} \\
    GPTQ + ReQuant    & \textbf{0.078} & \textbf{6.68}  & 0.101          & 3.86           & 0.105          & 3.04 \\
    \cmidrule(lr){1-7}
    GPTAQ             & 0.073          & \textbf{6.65}  & \textbf{0.116} & \textbf{3.94}  & \textbf{0.109} & \textbf{3.05} \\
    GPTAQ + ReQuant   & \textbf{0.073} & \textbf{6.65}  & 0.123          & 3.97           & 0.114          & 3.07 \\
    \bottomrule
  \end{tabular}
\end{table}

\begin{table}[ht]
  \centering
  \caption{Zero-shot accuracy (\%) under W4A16 on Llama-3 8B \emph{without} QuaRot rotation. \textbf{Avg.} is the mean over ten tasks.}
  \label{tab:requant-w4a16-acc-norot}
  \footnotesize
  \setlength{\tabcolsep}{2.5pt}
  \resizebox{\textwidth}{!}{%
  \begin{tabular}{@{}l|cccccccccc|c@{}}
    \toprule
    \textbf{Method} & \textbf{Arc C} & \textbf{Arc E} & \textbf{BoolQ} & \textbf{CEval} & \textbf{HellaSwag} & \textbf{LAMBADA} & \textbf{OBQA} & \textbf{PIQA} & \textbf{SIQA} & \textbf{Winogrande} & \textbf{Avg.$\uparrow$} \\
    \midrule
    FP                & 53.58 & 77.69 & 81.25 & 48.44 & 79.17 & 75.70 & 45.0 & 80.90 & 46.88 & 73.24 & 66.19 \\
    \midrule
    RTN               & 48.98 & 75.59 & 73.09 & 38.04 & 75.36 & 70.00 & 42.0 & 77.91 & \textbf{44.37} & 71.11 & 61.65 \\
    RTN + ReQuant     & \textbf{50.68} & \textbf{76.14} & \textbf{77.65} & \textbf{40.86} & \textbf{76.40} & \textbf{72.25} & \textbf{43.2} & \textbf{78.62} & 43.96 & \textbf{72.22} & \textbf{63.20} \\
    \cmidrule(lr){1-12}
    AWQ               & \textbf{50.85} & 73.82 & 78.53 & 43.24 & \textbf{77.60} & 70.29 & 43.0 & \textbf{79.71} & 46.11 & 71.90 & 63.50 \\
    AWQ + ReQuant     & 48.89 & \textbf{75.34} & \textbf{80.58} & \textbf{43.98} & 77.49 & \textbf{71.24} & \textbf{44.2} & 79.54 & \textbf{46.21} & \textbf{73.16} & \textbf{64.06} \\
    \cmidrule(lr){1-12}
    GPTQ              & 49.06 & 73.82 & \textbf{81.71} & \textbf{44.58} & 77.12 & \textbf{72.91} & 45.6 & \textbf{80.14} & \textbf{46.78} & 74.59 & \textbf{64.63} \\
    GPTQ + ReQuant    & \textbf{51.45} & \textbf{76.18} & 78.81 & 41.16 & \textbf{77.53} & 72.13 & \textbf{45.8} & 79.54 & 46.42 & \textbf{75.30} & 64.41 \\
    \cmidrule(lr){1-12}
    GPTAQ             & 50.43 & 74.75 & 79.36 & 42.50 & \textbf{77.81} & \textbf{71.82} & 45.2 & \textbf{79.49} & \textbf{46.21} & \textbf{75.22} & 64.28 \\
    GPTAQ + ReQuant   & \textbf{50.51} & \textbf{76.56} & \textbf{80.46} & \textbf{44.21} & 77.45 & 71.78 & \textbf{45.8} & 79.05 & 46.06 & 74.90 & \textbf{64.68} \\
    \bottomrule
  \end{tabular}}
\end{table}

\subsection{Effect of calibration sample size on Llama-3 8B at W4A4 (GPTAQ + QuaRot + ReQuant)}
\label{app:gptaq-calibration-samples}

\begin{figure}[t]
  \centering
  \includegraphics[width=0.32\textwidth]{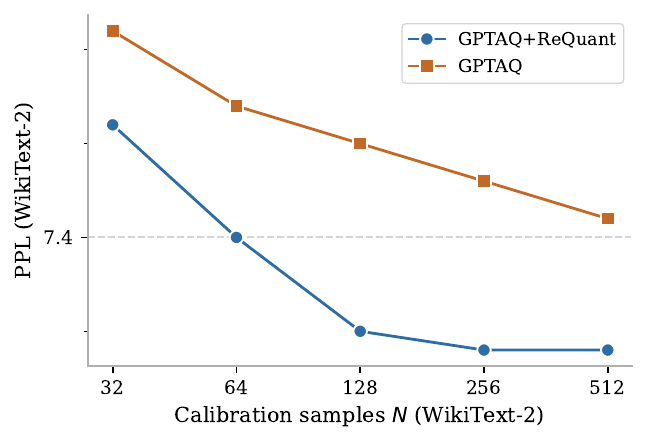}\hfill
  \includegraphics[width=0.32\textwidth]{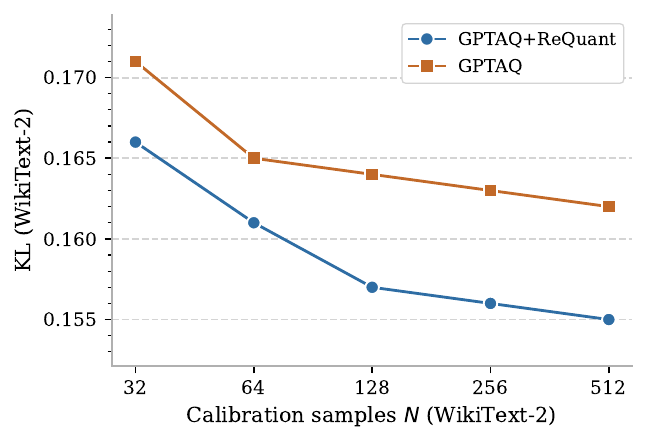}\hfill
  \includegraphics[width=0.32\textwidth]{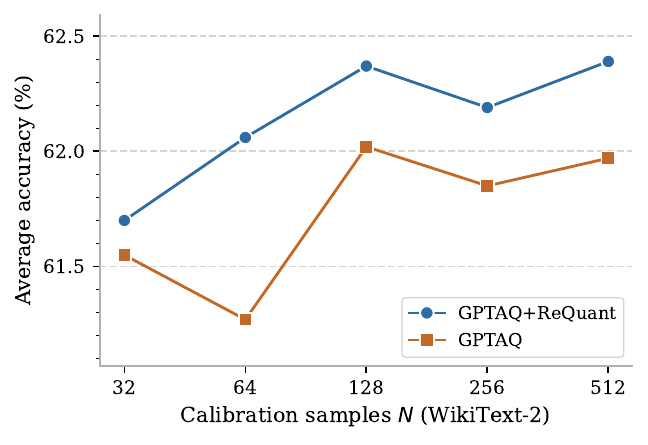}
  \caption{W4A4: sensitivity to calibration sample size for GPTAQ + QuaRot + ReQuant on Llama-3 8B. (left) WikiText-2 perplexity, (center) WikiText-2 KL divergence to the FP model, (right) average zero-shot accuracy (ten-task mean). Lower is better for PPL and KL; higher is better for accuracy.}
  \label{fig:gptaq-calibration-sweep}
\end{figure}

Figure~\ref{fig:gptaq-calibration-sweep} studies the effect of calibration-set size at \textbf{W4A4} on Llama-3 8B, with all other settings following Section~\ref{sec:exp-setup}. We report WikiText-2 perplexity, WikiText-2 KL divergence to the full-precision model, and average zero-shot accuracy from lm-eval-harness for GPTAQ~\cite{li2025gptaq} with QuaRot~\cite{ashkboos2024quarot}, both before and after ReQuant refinement. Increasing the number of calibration sequences generally improves all three metrics by providing more stable activation statistics, although the gains diminish once the calibration budget becomes sufficiently large.

Across all calibration-sample counts shown, ReQuant improves over GPTAQ+QuaRot under the same W4A4 deployment format: WikiText-2 perplexity and KL decrease while average zero-shot accuracy increases. The gains therefore appear across calibration budgets, not only under a particular sample count.

\subsection{Separating QuaRot from FlexRound comparisons}
\label{app:flexround-quarot}

Table~\ref{tab:flexround-quarot} reports controlled add-ons around the FlexRound comparison in Section~\ref{sec:flexround} on Llama-3 8B under W4A16.
Adding QuaRot to FlexRound improves PPL/KL and average accuracy, confirming that rotation is a beneficial and orthogonal pipeline component.
Pairing QuaRot with ReQuant further strengthens the RTN pipeline, and the paired GPTQ+QuaRot vs.\ GPTQ+QuaRot+ReQuant comparison isolates ReQuant under identical initialization, grid, scales, and zero-points: five of six PPL/KL metrics improve and average accuracy rises from $65.16\%$ to $65.33\%$.

\begin{table}[ht]
  \centering
  \caption{Controlled W4A16 comparisons on Llama-3 8B that separate QuaRot from ReQuant. GPTQ rows reuse Table~\ref{tab:t-cost-benefit}.}
  \label{tab:flexround-quarot}
  \footnotesize
  \setlength{\tabcolsep}{3.5pt}
  \resizebox{\linewidth}{!}{%
  \begin{tabular}{@{}lccccccc@{}}
    \toprule
    Method & Wiki PPL$\downarrow$ & Wiki KL$\downarrow$ & UC PPL$\downarrow$ & UC KL$\downarrow$ & NM PPL$\downarrow$ & NM KL$\downarrow$ & Acc.$\uparrow$ \\
    \midrule
    FlexRound & 6.69 & 0.075 & 3.73 & 0.067 & 2.97 & 0.081 & 65.08 \\
    FlexRound+QuaRot & 6.47 & 0.048 & \textbf{3.58} & 0.039 & 2.84 & 0.044 & 65.24 \\
    RTN+ReQuant ($T{=}4$, no QuaRot) & 7.63 & 0.174 & 3.94 & 0.116 & 3.05 & 0.100 & 61.78 \\
    RTN+QuaRot+ReQuant ($T{=}4$) & 6.52 & 0.055 & 3.62 & 0.042 & 2.85 & 0.045 & \textbf{65.42} \\
    GPTQ+QuaRot ($T{=}0$) & 6.53 & 0.054 & 3.60 & 0.038 & 2.85 & \textbf{0.042} & 65.16 \\
    GPTQ+QuaRot+ReQuant ($T{=}4$) & \textbf{6.43} & \textbf{0.041} & \textbf{3.58} & \textbf{0.033} & \textbf{2.83} & \textbf{0.042} & 65.33 \\
    \bottomrule
  \end{tabular}}
\end{table}

\subsection{Search-procedure ablations}
\label{app:search-ablations}

Unless noted, experiments below use Llama-3 8B, W4A16, GPTQ+QuaRot+ReQuant, $T{=}4$, and seed $0$.

\paragraph{Neighborhood size $K$.}
Candidate integer offsets are $\{\pm1,\ldots,\pm K\}$.
Table~\ref{tab:k-ablation} shows that $K{\in}\{1,2,3\}$ yields similar quality and runtime.
$K{=}2$ is strongest on WikiText PPL/KL for GPTQ and yields the highest RTN average accuracy, while $K{=}1$ is competitive on most PPL/KL metrics for RTN; we therefore use $K{=}2$ as the default.

\begin{table}[ht]
  \centering
  \caption{Neighborhood-size ablation at fixed $T{=}4$ on Llama-3 8B (W4A16).}
  \label{tab:k-ablation}
  \footnotesize
  \setlength{\tabcolsep}{4pt}
  \begin{tabular}{@{}llcccc@{}}
    \toprule
    Initializer & $K$ & Wiki PPL$\downarrow$ & Wiki KL$\downarrow$ & Acc.$\uparrow$ & Time (min)$\downarrow$ \\
    \midrule
    \multirow{3}{*}{RTN+QuaRot}
      & 1 & \textbf{6.52} & \textbf{0.054} & 64.84 & 80.72 \\
      & 2 & \textbf{6.52} & 0.055 & \textbf{65.48} & 80.89 \\
      & 3 & 6.53 & 0.055 & 65.08 & \textbf{80.19} \\
    \cmidrule(lr){1-6}
    \multirow{3}{*}{GPTQ+QuaRot}
      & 1 & \textbf{6.43} & 0.042 & 65.41 & \textbf{92.93} \\
      & 2 & \textbf{6.43} & \textbf{0.041} & 65.33 & 96.36 \\
      & 3 & \textbf{6.43} & \textbf{0.041} & \textbf{65.73} & 95.61 \\
    \bottomrule
  \end{tabular}
\end{table}

\paragraph{Coordinate ordering.}
Table~\ref{tab:order-ablation} compares forward ($j{=}1{\ldots}d_{\mathrm{col}}$), reverse, and a random-fixed within-row permutation reused across sweeps.
WikiText PPL differs by at most $0.0076$ and average accuracy by $0.37$ points.
Ordering changes the local solution while preserving overall quality; random-fixed is slower due to permutation overhead.

\begin{table}[ht]
  \centering
  \caption{Within-row coordinate-order ablation for GPTQ+QuaRot+ReQuant ($T{=}4$, $K{=}2$) on Llama-3 8B.}
  \label{tab:order-ablation}
  \footnotesize
  \begin{tabular}{@{}lcccc@{}}
    \toprule
    Order & Wiki PPL$\downarrow$ & Wiki KL$\downarrow$ & Acc.$\uparrow$ & Time (min)$\downarrow$ \\
    \midrule
    Forward & \textbf{6.43} & \textbf{0.041} & 65.33 & 96.36 \\
    Reverse & \textbf{6.43} & 0.042 & \textbf{65.70} & \textbf{96.06} \\
    Random-fixed & \textbf{6.43} & 0.042 & 65.60 & 122.65 \\
    \bottomrule
  \end{tabular}
\end{table}

\paragraph{QuaRot seed stability.}
Given fixed calibration data, ReQuant itself is deterministic; randomness enters through QuaRot.
Over five QuaRot seeds with GPTQ+QuaRot+ReQuant ($T{=}4$, $K{=}2$), WikiText PPL is $6.4277{\pm}0.0040$, WikiText KL is $0.0414{\pm}0.0004$, and average accuracy is $65.27{\pm}0.55$ (range $64.68$--$66.09$).
PPL/KL are stable across seeds, while downstream accuracy exhibits moderate seed variability.

\end{document}